\pdfoutput=1

\documentclass[11pt]{article}

\usepackage[final]{acl}

\usepackage{times}
\usepackage{latexsym}

\usepackage[T1]{fontenc}

\usepackage[utf8]{inputenc}

\usepackage{microtype}

\usepackage{inconsolata}

\usepackage{url}
\usepackage[utf8]{inputenc} 
\usepackage[T1]{fontenc}    
\usepackage{url}            
\usepackage{booktabs}       
\usepackage{amsfonts}       
\usepackage{nicefrac}       
\usepackage{microtype}      
\usepackage{xcolor}         

\usepackage{times}
\usepackage{soul}
\usepackage{url}
\usepackage{graphicx}
\usepackage{amsmath}
\usepackage{amsthm}
\usepackage{booktabs}
\usepackage{algorithm}
\usepackage[switch]{lineno}

\usepackage{amssymb}
\usepackage{pifont}
\usepackage{booktabs}
\usepackage{array}
\usepackage{algorithm}
\usepackage[noend]{algpseudocode}

\usepackage{wrapfig}
\usepackage{tabularx} 
\newcolumntype{Y}{>{\centering\arraybackslash}X}
\usepackage{threeparttable}
\usepackage{caption}
\usepackage{mathrsfs}
\usepackage{multirow}
\usepackage{bigstrut}
\usepackage{tcolorbox}

\usepackage{xcolor}
\definecolor{deepgreen}{rgb}{0.0, 0.5, 0.0}
\definecolor{deeperyellow}{rgb}{1, 0.8, 0}

\usepackage{amssymb}

\title{Can LLMs Evaluate Complex Attribution in QA? Automatic Benchmarking using Knowledge Graphs}

\author{
Nan Hu$^{1,2}$\thanks{Email: nanhu@seu.edu.cn},
Jiaoyan Chen$^{3}$\thanks{Corresponding authors. Email: gqi@seu.edu.cn, jiaoyan.chen@manchester.ac.uk, j.z.pan@ed.ac.uk},
Yike Wu$^{1,2}$,
Guilin Qi$^{1,6}$\footnotemark[2], \\
\bfseries Hongru Wang$^{7}$,
Sheng Bi$^{5}$,
Yongrui Chen$^{1,2}$,
Tongtong Wu$^{8}$,
Jeff Z. Pan$^{4}$\footnotemark[2]\\[4pt]
$^{1}$ School of Computer Science and Engineering, Southeast University\\
$^{2}$ Key Laboratory of New Generation Artificial Intelligence Technology and \\ 
its Interdisciplinary Applications
(Southeast University), Ministry of Education \\
$^{3}$ The University of Manchester
$^{4}$ The University of Edinburgh \\
$^{5}$ Law and Innovation Lab, School of Law, Southeast University \\ 
$^{6}$ State Key Lab. for Novel Software Technology, Nanjing University\\ 
$^{7}$ The Chinese University of Hong Kong 
$^{8}$ Monash University  \\
}

\begin{document}
\maketitle
\begin{abstract}
Attributed Question Answering (AQA) has attracted wide attention, but there are still several limitations in evaluating the attributions, including lacking fine-grained attribution categories, relying on manual annotations, and failing to compare attributions with only subtle differences.  
To bridge these gaps, we introduce Complex Attributed Question Answering (\textbf{CAQA}), a large-scale benchmark containing comprehensive attribution categories, automatically generated using Knowledge Graphs (KGs),  and complex attribution scenarios.
We have conducted extensive experiments to verify the effectiveness of CAQA, including the benchmarking of 25 automatic evaluators, their comparison with human evaluators, the testing of LLM evaluators fine-tuned by CAQA and so on.
These experiments also lead to a series of important findings that can benefit the future research of AQA.
All the codes and data are publicly accessible at \url{https://github.com/HuuuNan/CAQA-Benchmark}.
\end{abstract}

\section{Introduction}

Generative AI is increasingly adept together with other techniques like search engines to produce answers to natural language questions. However, their tendency to generate confident yet inaccurate or ``hallucinated'' contents \citep{ji2023survey,PRKS+2023} poses significant risks in high-stakes domains such as medicine \citep{lee2023benefits}. Therefore, Question Answering (QA) with attribution has been proposed, where not only answers but also citations (or evidence snippets) for supporting the answers are output \citep{menick2022teaching,rashkin2023measuring,bohnet2022attributed,li2023survey}.

Despite their potential, state-of-the-art implementations of Attributed QA (AQA), exemplified by Large Language Models (LLMs) with search engines like Bing Chat, perplexity.ai and YouChat\footnote{bing.com/new, perplexity.ai, https://you.com/}, still often produce erroneous attributions \citep{liu2023evaluating}. To develop systems for higher-quality attributions, it is crucial to explore effective automatic attribution evaluation methods, which can not only compare and analyze different AQA systems, but also provide feedback to improve their attributions \citep{asai2023self,yue2023automatic,gao2023rarr,bohnet2022attributed}, alleviating the issues of factuality, faithfulness and hallucination \citep{amouyal2022qampari,asai2023self}. 
However, the existing AQA benchmarks (see Table \ref{table1}) are inadequate due to their limited sizes, incomplete attribution categories (e.g., they do not fully consider \texttt{partially supportive} where only some sub-facts in the answer are supported), and ignore of complex attribution scenarios which require reasoning with multiple evidence under various logic operations, and are common in Bing Chat and retrieve-and-read systems \citep{malaviya2023expertqa}.

In this work, we adopt a comprehensive set of attribution categories including 
\texttt{supportive}, \texttt{partially supportive}, \texttt{contradictory} and \texttt{irrelevant} (see Table \ref{tab:CategoryExamples} for examples)
and define different levels of attribution complexity based on the reasoning required to infer the answer: \texttt{single}, \texttt{union}, \texttt{intersection}, and \texttt{concatenation} (see Table \ref{tab:examples_of_complexity} for examples).
Based on these foundations, we construct a larger-scale Complex Attributed Question Answering (CAQA) benchmark 
by an automatic generation method based on a Knowledge Graph (KG)~\cite{PVGW2017,PCEH+2017} composed of relational facts \citep{hogan2021knowledge,bollacker2008freebase} and two existing KGQA datasets containing question-answer pairs and corresponding KG queries.
%
Briefly, the construction method first extends a query by introducing additional logical operators for increasing reasoning complexity, then employs the extended queries to extract KG sub-graphs~\cite{HZWV+2024}, and finally edits these sub-graphs using different strategies and re-writes them into natural language citations as attributions of different categories. 
This method is flexible, allowing the generation of benchmarks with varied features, and adaptable to different KGs and KGQA datasets.

\begin{table}[!t]
\centering
\tiny
\caption{
CAQA and existing benchmarks. Attribution categories include Supportive (S), Non-supportive (N), Partially Supportive (P), Contradictory (C), Irrelevant (I) and Extrapolatory (E), with E and I treated as equivalent. Comp. denotes whether the benchmark considers attribution complexity. Auto. denotes whether the benchmark is automatically constructed without manual annotation.
}
\label{table1}
\begin{tabular}{lcccc}
\toprule
\textbf{Benchmarks} & \textbf{\#Sample}  & \textbf{Category} & \textbf{Comp.} & \textbf{Auto.}\\
\midrule
Bohnet et al. \citep{bohnet2022attributed} & 23,000    & \textsc{S/N} & \ding{55} & \ding{55} \\
{HAGRID} \citep{kamalloo2023hagrid} & 2,638    & \textsc{S/N} & \ding{55} & \ding{55} \\
{ExpertQA} \citep{malaviya2023expertqa} & 2,177   & \textsc{S/N} & \ding{55} & \ding{55} \\
{AttributionBench} \citep{li2024attributionbench} & 17,816    & \textsc{S/N}  & \ding{55} &  \ding{55}\\ 
Liu et al. \citep{liu2023evaluating} & 11,037    & \textsc{S/P/N} & \ding{55} & \ding{55} \\
{ALCE} \citep{gao2023enabling} & 800    & \textsc{S/P/N} & \ding{55} & \ding{55} \\
AttrEval-Gen \citep{yue2023automatic} & 242 & \textsc{S/C/E} & \ding{55} & \ding{55}\\
\midrule
{AttrEval-Sim} \citep{yue2023automatic} & 64.2K    & 
\textsc{S/C/E} & \ding{55} & \ding{51} \\ 
{CAQA (Ours)} & {161.1K}    & \textsc{S/P/C/I} &  \ding{51} & \ding{51} \\
\bottomrule
\end{tabular}

\end{table}

We use two existing attribution evaluators that are fine-tuned on specific data and 23 LLM evaluators under the zero-shot, few-shot and fine-tuning settings to demonstrate the effectiveness of CAQA for assessing and developing AQA evaluators. Through extensive evaluation, we get a series of important findings. 
(1) All the evaluators struggle to identify the nuanced negative attribution categories in both zero-shot and few-shot settings. 
With fine-tuning, the F1 scores of all the categories exceed 90\% for most LLM evaluators.
(2) All the evaluators perform poorly in recognizing attributions of \texttt{partially supportive} and perform worse on more complex attribution scenarios, e.g. the GPT models perform worse on those requiring \texttt{concatenation} and \texttt{intersection}, while other open source LLMs perform worse on those requiring \texttt{union}.
(3) The automatically generated CAQA is reliable with high consistency as human annotators.
(4) When tested on an out-of-distribution dataset, LLM evaluators fine-tuned by CAQA outperform the existing particularly developed evaluators.


\section{Related Work}

\noindent\textbf{Attributed Question Answering}.
LLMs now lead the performance in QA, but often produce hallucinations \citep{ji2023survey,xiao-wang-2021-hallucination,wang-sennrich-2020-exposure,shuster-etal-2021-retrieval-augmentation}. To alleviate this issue, some studies \citep{menick2022teaching,nakano2021webgpt,gao2023enabling} train attributed models to answer questions that generate answers along with supporting evidence, typically in the form of citations or references.
Some other studies augment LLMs with external tools \citep{mialon2023augmented,shen2023hugginggpt,schick2023toolformer,WQLPW2024,wang2025rethinking,HNWY+2025} such as retrievers \citep{han-etal-2023-improving,shi2023replug,asai2023self,izacard2022few,DBLP:conf/naacl/MinHJLCCLQLLW24,DBLP:conf/naacl/MinXQHY25,SDVM+2025} and search engines \citep{nakano2021webgpt,komeili2021internet}, or incorporate external references for attribution.
 While these approaches improve factual grounding, ensuring attribution accuracy and consistency remains an open challenge, underscoring the need for more systematic evaluation.

Attributed QA is conceptually related to the task of Attributing Unanswerable Questions (AUQ) \citep{DBLP:conf/sigir-ap/MoradisaniZSN24}, although the two tasks differ in scope. AUQ focuses on identifying questions that cannot be answered based on the given context, typically due to factors such as negation, entity swaps, or missing information, and explaining why a valid answer does not exist. In contrast, AQA aims to generate answers along with explicit attribution from the context. A well-designed AQA system should naturally refrain from answering when no supporting evidence is available, thereby addressing unanswerability through its attribution mechanism.

\noindent\textbf{Attribution Evaluation}.
Current methods for evaluating attribution predominantly depend on human annotation \citep{nakano2021webgpt,bohnet2022attributed,liu2023evaluating,rashkin2023measuring,muller2023evaluating}, which is costly and inefficient. 
Recent studies propose automatic attribution evaluators based on LLMs, such as \textsc{AutoIS} \citep{gao2023rarr,bohnet2022attributed} and \textsc{AttrScore} \citep{yue2023automatic}. 
However, existing benchmarks are inadequate for evaluating and advancing attribution evaluators due to their limited sizes and restricted evaluation settings, including incomplete attribution categories and the ignorance of attribution complexity. They mostly classify attribution into only two categories: the cited evidence \textit{supports} or \textit{does not support} the answer \citep{gao2023enabling,li2023towards,li2024attributionbench,malaviya2023expertqa,bohnet2022attributed}. 
Some benchmarks \citep{gao2023enabling,liu2023evaluating, ZhangAYPHK24} add a third category, \textit{partially supportive}, but their sizes are small and reliance on manual annotation. 
\cite{yue2023automatic} presents a method for automatically generating attribution annotations to construct large-scale samples with categories of \textit{supportive}, \textit{contradictory}, and \textit{extrapolatory} (equivalent to \textit{irrelevant}).
However, their method cannot support \textit{partially supportive}, as it relies solely on answer word replacement.
Our work addresses these limitations by proposing a novel method based on KGs and KGQA datasets to automatically create a large-scale AQA benchmark with comprehensive attribution categories. Notably, our benchmark is the first to offer fine-grained evaluation for partially supportive evidence and consider varying levels of logical reasoning complexity in attribution.

\section{Definitions in QA Attribution}
\label{section3}

\subsection{Task Formulation}


This work aims to evaluate the attribution in AQA. It is to verify whether an evidence, which has one or multiple citations (references) with facts stated, can sufficiently support a generated answer towards a natural language question. 
%
Formally, given a question $q$, an answer statement $a$ and an evidence $e$, the objective of attribution evaluation is to map them to an attribution category $t$ (a.k.a. class label).
It can be represented by the function $\mathcal{F}:\mathcal{Q}\times\mathcal{A}\times\mathcal{E}\mapsto\mathcal{T}$, where $\mathcal{Q}$, $\mathcal{A}$ and $\mathcal{E}$ denote the sets of questions, answers and evidence, respectively,
and $\mathcal{T}$ denotes the set of potential categories, such as $\{$\textit{supportive}, \textit{partially supportive}, \textit{contradictory}, \textit{irrelevant}$\}$ which mean ``$e$ is supportive, partially supportive, contradictory or irrelevant to the fact that $a$ is the answer of $q$.''

\subsection{Attribution Categorization}

\begin{table*}[!t]
\centering
\small
\caption{Examples of the four attribution categories. \textcolor{deepgreen}{Green}, \textcolor{deeperyellow}{yellow}, and \textcolor{red}{red} text indicate the content in the answer that is \textcolor{deepgreen}{supported}, \textcolor{deeperyellow}{not supported}, or \textcolor{red}{contradicted} by the content in the citation, respectively.}
\label{tab:CategoryExamples}
\scalebox{0.9}{
\begin{tabularx}{1.1\textwidth}{lX}
\toprule
\textbf{Attribution Category} & \textbf{Examples}  \\
\midrule
\raisebox{-2.9\height}{\textbf{Supportive}} & \textbf{Question:} Who plays Fruma Sarah in Fiddler on the Roof?
\newline 
\textbf{Answer:} \textcolor{deepgreen}{Fruma Sarah is a character in the musical ``Fiddler on the Roof'}', and \textcolor{deepgreen}{Ruth Madoc played the role} [1].
\newline \textbf{Citations:} [1] ... In 1971 \textcolor{deepgreen}{Ruth Madoc played Fruma Sarah in the film version of the musical ``Fiddler on the Roof''}, and in 1972 she appeared as ...
\\

\midrule

\raisebox{-3.2\height}{\textbf{Partially Supportive}} & \textbf{Question:} Who plays Patrick in 10 Things I Hate About You?
\newline \textbf{Answer:} \textcolor{deeperyellow}{Patrick is played by} \textcolor{deepgreen}{actor Heath Ledger in the film 10 Things I Hate About You} [1].
\newline \textbf{Citations:} [1] \textcolor{deepgreen}{10 Things I Hate About You} is a 1999 American teen romantic comedy-drama film directed by Gil Junger and \textcolor{deepgreen}{starring Heath Ledger}, Julia Stiles, Joseph Gordon-Levitt, and Larisa Oleynik. The screenplay, written by ... 
\\

\midrule

\raisebox{-2.4\height}{\textbf{Contradictory}}  & \textbf{Question:} Who directed a George Pal's production?
\newline \textbf{Answer:} \textcolor{red}{George Pal directed a production called Puppetoons} [1].
\newline \textbf{Citations:} [1] ... \textcolor{red}{The Puppetoon} Movie is a 1987 animated film written, produced, and \textcolor{red}{directed by Arnold Leibovit} ...
\\

\midrule

\raisebox{-3.2\height}{\textbf{Irrelevant}} & \textbf{Question:} Who played the weasley brothers in Harry Potter?
\newline \textbf{Answer:} \textcolor{deeperyellow}{James and Oliver Phelps, identical twin actors, played the roles of Fred and George Weasley in the Harry Potter film series} [1].
\newline \textbf{Citations:} [1] Chris Rankin plays of ``Bugsy Malone'', ``The Lion, The Witch and The Wardrobe'' and Harry Potter series ... he plays a brother of Harry Potter's best friend, ...
\\

\bottomrule
\end{tabularx}
}
\end{table*}

We analyse the results of practical AQA systems \citep{gao2023enabling} and find that apart from correct attributions that are \textit{supportive}, there are three main categories of incorrect attributions: \textit{partially supportive}, \textit{contradictory} and \textit{irrelevant}.
More details are shown in Appendix \ref{appendix:AnalysiOfAQASytems}.
The four attribution categories are defined below:

\noindent $\bullet$ \textbf{Supportive (Sup.)}: The evidence includes facts that can fully support the answer statement.

\noindent $\bullet$ \textbf{Partially Supportive (Par.)}: The evidence lacks a part of the facts that are required to infer the answer statement.

\noindent $\bullet$ \textbf{Contradictory (Con.)}: The evidence includes facts that can infer a different answer statement.

\noindent $\bullet$ \textbf{Irrelevant (Irr.)}: The evidence has no facts that can be used to infer the answer statement. 

Table \ref{tab:CategoryExamples} provides examples of the four attribution categories. In the \textbf{supportive} example, the answer is backed by citation [1], which confirms that “\textit{Ruth Madoc plays Fruma Sarah in Fiddler on the Roof}.” In the \textbf{partially supportive} example, the answer cites [1] but does not fully align with the complete context provided, mentioning only “\textit{the actor Heath Ledger stars in the film 10 Things I Hate About You}” and missing the information ``\textit{Heath Ledger plays the character Patrick}''. 
In the \textbf{contradictory} example, the citation [1] states “\textit{The Puppetoon Movie is directed by Arnold Leibovit},” which contradicts the generated answer. The \textbf{irrelevant} example involves citing [1], which discusses an unrelated actor, Chris Rankin, and his career offers no relevant facts to verify the answer.
Note that the contradictory category differs from the partially supportive or irrelevant categories in reasoning. In the former, the evidence leads to another answer that conflicts with the generated answer.



\subsection{Attribution Complexity}

Previous research has not explored different levels of complexity in inferring the answer. 
\citet{malaviya2023expertqa} has shown that AutoIS \citep{bohnet2022attributed}, the most commonly used automatic attribution evaluator, often mistakes in scenarios that require multiple pieces of evidence to validate the answer.
To advance automatic evaluators, our benchmark incorporates reasoning complexity by categorizing attribution into four levels of complexity, based on the reasoning logic of supporting facts in the citations  (see Table \ref{tab:examples_of_complexity} for examples):
%

\noindent $\bullet$ \textbf{Single}: The answer is supported by one single citation.

\noindent $\bullet$ \textbf{Union}: The answer is supported by independent facts from multiple citations.
  
\noindent $\bullet$ \textbf{Intersection}: The answer is supported by facts with common entities from multiple citations.

\noindent $\bullet$ \textbf{Concatenation}: The answer is supported by chains of facts from multiple citations.
  


\section{Benchmark Construction}
\label{section4}

\begin{figure*}[!t]
  \centering
  \includegraphics[width=0.85\textwidth]{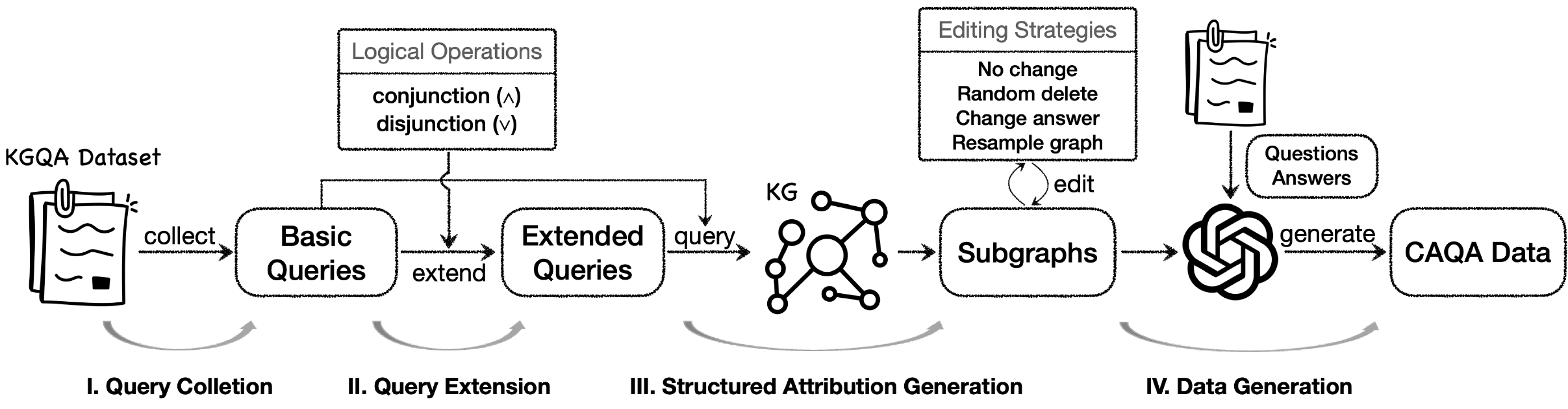}
  \caption{The entire process of constructing the CAQA benchmark.
  } 
  \label{fig:Framework}

\end{figure*}

In this section, we introduce our benchmark construction method. Figure \ref{fig:Framework} presents its overview including four key steps:
(1) Query Collection: given a KGQA dataset, we collect data corresponding to basic KG logical queries;
(2) Query Extension: two logical operators are applied to increase the complexity of the original queries; (3) Structured Attribution Generation: the extended queries are grounded in the KG to obtain relevant subgraphs, which are then probabilistically edited using four strategies to generate new subgraphs with four attribution categories; (4) Data Generation: we produce AQA data, where each instance consists of an extended question, rephrased answer entities, citations derived from subgraphs, as well as the attribution category and complexity labels.

\vspace{-0.1cm}
\subsection{Query Collection}
The selection of KGs and KGQA datasets is primarily motivated by two observations: (1) KGQA is a well-established task with a wealth of open resources, as evidenced by 25 KGQA datasets for 5 KGs reported in \citep{jiang2022knowledge}; 
(2) existing KGQA datasets contain high-quality question-answer pairs and corresponding KG logical queries, often expressed in SPARQL, which are capable of deriving the correct answers and can be leveraged to generate evidence.

A KG is composed of relational facts in the form of triple, i.e., $(h, r, t)$, where $h$ and $t$ denote a subject entity and a object entity, respectively, and $r$ denotes a relation between them.
A KGQA dataset $D = \{S_1, S_2, ..., S_N\}$ consists of samples in the form of $ S_i = (q_i, a_i, l_i)$, where $q_i$ denotes a natural language question, $a_i$ denotes its answer entity, and $l_i$ denotes the KG logical query of $q_i$. Our data collection focuses on samples where the KG logical query falls into one of three types: \textbf{single-triple}, \textbf{path-like}, or \textbf{tree-like queries}.
As shown in the first three columns in Table \ref{table3}, a single triple query denoted as $(e_0, r_0, ?a)$ indicates that the answer entity $?a$ can be obtained via the subject $e_0$ and the relation $r_0$. 
A path-like query denoted as $[e_0, r_0, ?v_1, \ldots, ?v_{n-1}, r_{n-1}, ?a]$ represents that the answer $?a$ is reachable through an $n$-hop path starting from $e_0$, traversing $n$ relations and $n-1$ intermediate entities represented by query variables $?v_1, \ldots, ?v_{n-1}$. Specifically, $?v_1$ denotes the first intermediate variable in the query path. Notably, a path-like query reduces to a single-triple query when $n=1$.
A tree-like query, formulated as $\wedge_{i=0}^{n-1}(e_i, r_i, ?a)$, includes $n$ distinct triples, each originating from different subjects and converging on the answer $?a$.


\subsection{Query Extension}

\begin{table*}
\centering
\caption{The rules for extending the logical queries utilizing two query operations: intersection ($\wedge$) and union ($\vee$). All queries are classified according to their structure as single-triple (S.) queries, path-like (P.) queries, tree-like (T.) queries and union-tree-like (U.) queries. The `Examples' column presents corresponding graph representations for the case where $n=2$, $m=2$, and $k=0$, where grey nodes represent variables for answer entities.
}
\label{table3}
\resizebox{\textwidth}{!}{
\renewcommand{\arraystretch}{1.5}
\begin{tabular}{ccc|ccc}
\toprule
\hline
\multicolumn{3}{c|}{\textbf{Original Query ${l}$}}  &\multicolumn{3}{c}{\textbf{Extended Query ${l}^{\prime}$}}  \\

\cmidrule{1-3}
\cmidrule{4-6}

{\textbf{Definitions}} &{\textbf{Structures}} & \textbf{Examples} & 
{\textbf{Definitions}} & {\textbf{Structures}} & \textbf{Examples} \\ 


\hline

\multirow{2}{*}{($e_0, r_0, ?a$)} & \multirow{2}{*}{\textsc{S.}} & \multirow{2}{*}{\includegraphics[scale=0.5]{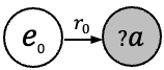}} &
\raisebox{-0.3\height}{($e_0, r_0, ?a) $} & \multirow{2}{*}{U.} & \multirow{2}{*}{\includegraphics[scale=0.5]{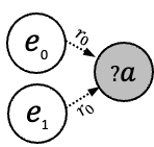}} \\
& &  & \raisebox{0.3\height}{$\vee (e_1, r_0, ?a) \vee \ldots \vee (e_m, r_0, ?a) $} & & \\

\midrule

\multirow{4}{*}{$[e_0, r_0, ?v_1, \ldots, ?v_{n-1}, r_{n-1}, ?a]$ } & \multirow{4}{*}{P.} & \multirow{4}{*}{\includegraphics[scale=0.5]{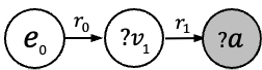}}  & 
\raisebox{-0.1\height}{$[e_0, r_0, ?v_1, \ldots, ?v_{n-1}, r_{n-1}, ?a]$} & \multirow{2}{*}{\raisebox{-0.\height}{P.}} & \multirow{2}{*}{\includegraphics[scale=0.5]{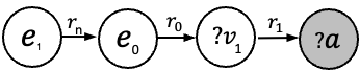}} \\

& & &  \raisebox{0.5\height}{$\land (e_{1}, r_{n}, e_0)$} & &\\
\cmidrule{4-6}

& & & \raisebox{-0.0\height}{$[e_0, r_0, ?v_1, \ldots, ?v_{n-1}, r_{n-1}, ?a]$} & 
\raisebox{-0.8\height}{T.} & \multirow{2}{*}{\includegraphics[scale=0.5]{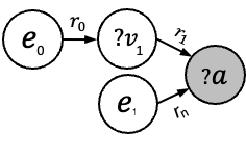}}  \\
& & & \raisebox{1.0\height}{$\land (e_{1}, r_{n}, ?a )$} & & \\

\midrule
\multirow{4}{*}{$\wedge_{i=0}^{n-1}(e_i, r_i, ?a) 
$} & \multirow{4}{*}{T.}  & \multirow{4}{*}{\includegraphics[scale=0.5]{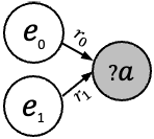}} &
\raisebox{-0.1\height}{$\wedge_{i=0}^{n-1}(e_i, r_i, ?a), \  i \neq k $ }& \multirow{2}{*}{T.} & \multirow{2}{*}{\includegraphics[scale=0.5]{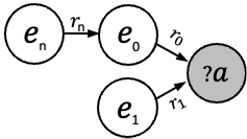}} \\ [1pt]

&  &  &  \raisebox{0.4\height}{$ \land (e_{n}, r_{n}, e_k) \land (e_k, r_k, ?a) $} &  \\
\cmidrule{4-6}

& &  & 
\multirow{2}{*}{$\wedge_{i=0}^{n-1}(e_i, r_i, ?a) \land (e_{n}, r_{n}, ?a)$}  & \multirow{2}{*}{T.} & \multirow{2}{*}{\includegraphics[scale=0.5]{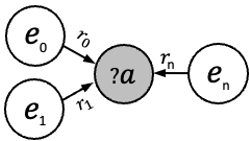}}  \\

&  &  &   &  &\\

\bottomrule

\end{tabular}
}

\vspace{-0.1cm}

\end{table*}

For each KGQA  example $S_i = (q_i, a_i, l_i)$, we extend the logical query $l_i$ to $l_i'$ using a set of predefined query extension rules  designed based on the logical operations \textit{intersection} (i.e., conjunction, $\wedge$) and \textit{union} (i.e., disjunction, $\vee$) \citep{ren2023neural}\footnote{Our method can be easily extended  with more logical operations like Negation and Kleene Plus \citep{ren2023neural}.
}.

Table \ref{table3} outlines the extension rules and we show concrete examples of each extension rule in Appendix \ref{appendix:ConcreteCasesofQueryExtension}.
For a single-triple query $l$, the \textit{union} operation is used.
Initially, we retrieve entities from the KG that share the same name as $e_0$ in $l$, producing a set of $m$ entities $\{ e_1, \ldots, e_m \}$, where $m$ may be zero.
Subsequently, we generate logical queries $(e_1, r_0, ?a)$, $\ldots$, $(e_m, r_0, ?a)$ by combining the retrieved entities and the relation $r_0$ from $l$. These new queries are then merged with $l$ via \textit{union}, resulting in a union-tree-like query structure. This structure implies that the final answer is derived as the union of the answers obtained from each subquery. 

For a path-like query or a tree-like query, we apply the \textit{intersection} operation in two distinct ways.
In the first way, we identify a unique subject entity $e_0$ for path-like queries or randomly select a subject $e_k$ for tree-like queries. We then retrieve corresponding triples $(e_1, r_{n}, e_{0})$ or $(e_n, r_{n}, e_{k})$ from the KG, where $r_{n}$ represents a relation not present in $l$. These new triples are appended to the respective queries, ensuring that $e_0$ and $e_k$ are connected entities.
This process maintains the overall structure of the path-like or tree-like query. 
In the second way, we append a new query $(e_{1},r_{n} ,?a)$ or $(e_{n},r_{n} ,?a)$ to the respective logical forms, ensuring that the intersection of the answers obtained from the new queries with those from $l$ is non-empty. Through this extension, both the path-like query and tree-like query are converted into the tree-like structures.

For both a path-like query ($n \geq 2$) and a tree-like query, the two intersection extensions are applied with equal probability. In contrast, for single-triple queries (a special case of path-like queries), four operations are equally likely: union extension, two types of intersection extension, and no extension (to preserve some single-triple queries).
The extension process results in four query types: \textit{single-tree}, \textit{union-tree-like}, \textit{tree-like}, and \textit{path-like}, corresponding to the attribution complexity types (denoted by $r$)—\textit{single}, \textit{union}, \textit{intersection}, and \textit{concatenation}.

\vspace{-0.15cm}
\subsection{Structured Attribution Generation}
\label{sec:GenerateAttribution}
\vspace{-0.05cm}

We first obtain a KG subgraph $\mathcal{G}$ by grounding each extended query ${l}^{'}$ in the KG, which returns the entities that are assigned to all the variables in the query for inferring the answer. The subgraph $\mathcal{G}$ is regarded as the structured attribution to support the answer to the question and falls under the \textit{supportive} attribution category.
To get structured attributions of \textit{partially supportive}, \textit{contradictory}, and \textit{irrelevant}, we edit  $\mathcal{G}$ as follows.

\noindent $\bullet$  \textbf{Partially Supportive}. The \textit{partially supportive} subgraph $\mathcal{G}_{In}$ is generated by partial deletion, resulting in a subgraph that cannot fully support the answer.  For path-like queries, we randomly delete one triple in $\mathcal{G}$. For tree-like or union-tree queries, we delete a path connecting one of the subject entities to the answer. In the case of single-triple queries, no deletion is performed.

\noindent $\bullet$ \textbf{Contradictory} The \textit{contradictory} subgraph $\mathcal{G}_{C}$ is constructed by altering $\mathcal{G}$ such that its reasoning conflicts with the answer. This is done by replacing the answer entity in $\mathcal{G}$ with a non-answer entity of the same type. Especially for queries involving a union operation, we replace one of the answer entities within $\mathcal{G}$.

\noindent $\bullet$ \textbf{Irrelevant} The \textit{irrelevant} subgraph $\mathcal{G}_{Ir}$ is obtained by selecting an entirely different subgraph from the KG that is structurally similar to $\mathcal{G}$ but contains unrelated entities and relations, except for the subject entity in $\mathcal{G}$.

\vspace{-0.15cm}
\subsection{Data Generation}
\vspace{-0.05cm}


We employ ChatGPT with tailored prompts to transform the subgraphs of $\mathcal{G}$, $\mathcal{G}_{In}$, $\mathcal{G}_{C}$ and $\mathcal{G}_{Ir}$ into citations (attributions) of  \textit{supportive}, \textit{partially supportive}, \textit{contradictory} and \textit{irrelevant}, respectively.
When the original logical query $l$ is expanded to $l'$, the original question $q$ is similarly extended to a new question $\tilde{q}$ using ChatGPT. In addition, the answer entity $a$ is paraphrased into a more detailed answer statement $\tilde{a}$.
Ultimately, it yields an AQA sample consisting of the question $q$ or $\tilde{q}$, the answer statement $\tilde{a}$, the citation ${c}$, the attribution category $t$, and the complexity label $r$. More details on this step can be found in Appendix \ref{appendix:GerationOfQAA}.



\section{Experiment Setup}




\begin{table}
\centering
\caption{Statistics of the CAQA dataset}
\label{table4}
\scalebox{0.7}{
\begin{tabular}{@{}lcccc@{}}
\toprule
\multicolumn{2}{l}{\multirow{2}{*}{\textbf{Classes}}} & \textbf{Train}                       &  \textbf{Test}                       & \textbf{Total}                       \\ 
\cmidrule(l){3-3} 
\cmidrule(l){4-4} 
\cmidrule(l){5-5} 
\multicolumn{2}{l}{}                                    & \multicolumn{1}{l}{137,211} & \multicolumn{1}{l}{23,963} & \multicolumn{1}{l}{161,174} \\ \midrule
\multirow{4}{*}{Category}     & \multicolumn{1}{c|}{Sup.} & 39,489                      & 6,668                      &               46,157               \\
                            & \multicolumn{1}{c|}{Par.}  & 28,868                      & 5,065                      &                      33,933      \\
                            & \multicolumn{1}{c|}{Con.} & 36,620                      & 6,423                      &                       43,043       \\
                            & \multicolumn{1}{c|}{Irr.}   & 32,234                      & 5,807                      &                     38,041       \\ \midrule
\multirow{4}{*}{Complexity} & \multicolumn{1}{c|}{Single}   & 73,795                      & 10,443                      &                  84,238           \\
                            & \multicolumn{1}{c|}{Concatenation}   & 46,783                      & 8,455                      &                      55,238       \\ 

                            & \multicolumn{1}{c|}{Union}   & 5,347                       & 886                        &                     6,233        \\
                            & \multicolumn{1}{c|}{Intersection}   & 11,286                      & 4,179                      &                     15,465        \\

\bottomrule
\end{tabular}
}

\vspace{-0.3cm}

\end{table}

\noindent\textbf{The CAQA Benchmark}.
It is constructed using two KGQA datasets: {GrailQA} \citep{gu2021beyond} and {WebQuestionsSP} \citep{yih2016value}, and the Freebase KG \citep{bollacker2008freebase}. 
See Table \ref{table4} for its statistics.
Additionally, we manually annotated the attribution categories of 300 test samples to assess their consistency with the original categories (see results in Section \ref{sec:cbh}). 
More details on CAQA are in Appendix \ref{appendix:CAQAConstruction}, and human annotation processes are described in Appendix \ref{appendix:HumanAnnotation}.


\noindent\textbf{ALCE-FineGrained}.
We manually annotated 215 samples of the ALCE benchmark with the four attribution categories, and get a new benchmark, ALCE-FineGrained, as an out-of-distribution testing set (see results in Section \ref{sec:ExplorationOfOOD}).
Details of human annotation are given in Appendix \ref{appendix:HumanAnnotation}.


\noindent\textbf{Attribution Evaluators}.
We evaluate state-of-the-art LLMs as attribution evaluators across different architectures and  sizes: GPT-3.5, GPT-4o, GPT-4o mini, LLaMA-2/3/3.1 \cite{grattafiori2024llama3herdmodels}, Vicuna \citep{vicuna2023}, Gemma-2 \cite{gemmateam2024gemma2improvingopen}, Mistral \cite{jiang2023mistral}, Phi-3 \cite{abdin2024phi3technicalreporthighly}, and Qwen2.5 \cite{hui2024qwen25codertechnicalreport}. We conduct evaluations in three settings: the zero-shot setting, the few-shot setting where several attribution examples are given, and the fine-tuning setting where the LLM is trained with the samples in the training set. 
Additionally, we test two specially developed automatic attribution evaluators \textsc{AutoIS} \citep{honovich-etal-2022-true-evaluating} and \textsc{AttrScore} \citep{yue2023automatic}.
%
More details on the evaluators' implementation are given in Appendix \ref{appendix:ImplementationDetails}.

\noindent\textbf{Metrics}.
We report F1 score for the performance on each attribution category and micro-F1 score for the performance on each complexity level and overall performance. 
We also use the \textsc{FActScores} metric \citep{emnlp/MinKLLYKIZH23} for a fine-grained evaluation of the \texttt{partially supportive} category.

\vspace{-0.2cm}
\section{Evaluation}

\subsection{Overall Results}

\begin{table*}[!t]
  \centering
  \small
  \begin{minipage}[b]{1\textwidth} 
    \centering
\scalebox{0.8}{
\begin{tabular}{llccccc|cccc}
\toprule
\multirow{2}{*}{\textbf{Settings}} & \multirow{2}{*}{\textbf{Evaluators (LLM Size)}} & \multicolumn{5}{c}{\textbf{Category}}  & \multicolumn{4}{|c}{\textbf{Complexity}}    \\ 
\cmidrule{3-7}
\cmidrule{8-11}

& & \multicolumn{1}{c}{Sup.}       & \multicolumn{1}{c}{Par.}       & \multicolumn{1}{c}{Con.} & \multicolumn{1}{c}{Irr.}    & Overall      
 & \multicolumn{1}{c}{Single}       & \multicolumn{1}{c}{Concatenation}     & \multicolumn{1}{c}{Intersection}   & \multicolumn{1}{c}{Union}      \\ 
\midrule


& LLaMA-3 (8B)                          & \multicolumn{1}{c}{0.467}          & \multicolumn{1}{c}{0.120}           & \multicolumn{1}{c}{0.072}             & \multicolumn{1}{c}{0.007}             & 0.296     & \multicolumn{1}{c}{0.304}          & \multicolumn{1}{c}{0.271}          & \multicolumn{1}{c}{0.283}          & \multicolumn{1}{c}{0.259}     \\ 

& LLaMA-3.1 (8B)                  &  0.544&	0.049&	0.130&	0.017&	0.318&	0.319&	0.326&	0.319&	0.285     \\ 

& LLaMA-3.1 (70B)  & 0.688& 0.168& 0.547& 0.609& 0.544& 0.545& 0.549& 0.545& 0.499             \\

& Mistral-v0.3 (7B)    & 0.661& 0.160& 0.051& 0.334& 0.362& 0.363& 0.374& 0.356& 0.337   \\ 

& Mixtral-v1.0 (8x7B)    &  0.677& 0.094& 0.170& 0.635& 0.494& 0.495& 0.516& 0.484& 0.487 \\

 & Vicuna (7B)                           & \multicolumn{1}{c}{0.513}          & \multicolumn{1}{c}{0.100}           & \multicolumn{1}{c}{0.064}           & \multicolumn{1}{c}{0.199}           &   0.327       & \multicolumn{1}{c}{0.343}          & \multicolumn{1}{c}{0.273}          & \multicolumn{1}{c}{0.312}          & \multicolumn{1}{c}{0.256}    \\ 

&  Vicuna (13B)                          & \multicolumn{1}{c}{0.634}          & \multicolumn{1}{c}{0.211}           & \multicolumn{1}{c}{0.393}             & \multicolumn{1}{c}{0.275}             & 0.405  
 & \multicolumn{1}{c}{0.432}          & \multicolumn{1}{c}{0.314}          & \multicolumn{1}{c}{0.361}          & \multicolumn{1}{c}{0.374}  \\

\textbf{Zero-Shot}  & Gemma-2 (9B)    &   0.667& 0.280& 0.498& 0.624& 0.556& 0.557& 0.572& 0.552& 0.508\\

& Gemma-2 (27B)    &  0.653& 0.184& 0.569& 0.646& 0.566& 0.566& 0.579& \underline{0.566}& 0.537 \\

& Qwen-2.5 (14B)    &  0.680& 0.132& \underline{0.708}& \underline{0.660}& \underline{0.617}& 0.640& \textbf{0.622}& \textbf{0.611}& 0.547 \\

& Qwen-2.5 (72B)    &  0.629& 0.266& 0.701& 0.471& 0.571& 0.593& 0.583& 0.565& 0.530 \\

& GPT-4                                 & \multicolumn{1}{c}{\textbf{0.771}} & \multicolumn{1}{c}{\textbf{0.456}} & \multicolumn{1}{c}{\textbf{0.745}} & \multicolumn{1}{c}{{{0.473}}} & \textbf{0.630} 
& \multicolumn{1}{c}{\textbf{0.685}} & \multicolumn{1}{c}{{0.451}} & \multicolumn{1}{c}{{0.514}} & \multicolumn{1}{c}{\textbf{0.616}} \\ 

& GPT-4o   &  \underline{0.769}& \underline{0.445}& 0.598& 0.626& \textbf{0.630}& \underline{0.676}& \underline{0.591}& 0.470& \underline{0.588} \\

& GPT-4o-mini   & 0.718& 0.297& 0.632& \textbf{0.703}& {0.616}& 0.672& 0.473& 0.444& 0.559 \\

\midrule


& LLaMA-3 (8B)                         & \multicolumn{1}{c}{0.573}          & \multicolumn{1}{c}{0.202}           & \multicolumn{1}{c}{0.234}             & \multicolumn{1}{c}{0.156}             & 0.336       
 & \multicolumn{1}{c}{0.356}          & \multicolumn{1}{c}{0.279}          & \multicolumn{1}{c}{0.310}          & \multicolumn{1}{c}{0.294}                 \\ 

& LLaMA-3.1 (8B)  &  0.631&  0.101&  0.307&  0.157&  0.353& 0.354&  0.375&  0.343&  0.362           \\ 

& LLaMA-3.1 (70B)     & 0.698& 0.259& 0.713& 0.665&0.627& 0.626& \underline{0.638}& \underline{0.632}& \underline{0.589}         \\

& Mistral-v0.3 (7B)  & 0.571 & 0.072& 0.252& 0.134& 0.342& 0.343& 0.351& 0.336& 0.326  \\ 

& Mixtral-v1.0 (8x7B)  &  0.589& 0.084& 0.356 & 0.558& 0.455& 0.456& 0.473& 0.451& 0.416\\

&   Vicuna (7B)                           & \multicolumn{1}{c}{0.578}          & \multicolumn{1}{c}{0.183}           & \multicolumn{1}{c}{0.081}           & \multicolumn{1}{c}{0.324}           & 0.325     & \multicolumn{1}{c}{0.337}          & \multicolumn{1}{c}{0.272}          & \multicolumn{1}{c}{0.354}          & \multicolumn{1}{c}{0.311}               \\

 & Vicuna (13B)                           & \multicolumn{1}{c}{0.633}          & \multicolumn{1}{c}{0.208}           & \multicolumn{1}{c}{0.383}           & \multicolumn{1}{c}{0.288}           & 0.403        & \multicolumn{1}{c}{0.427}          & \multicolumn{1}{c}{0.315}          & \multicolumn{1}{c}{0.397}          & \multicolumn{1}{c}{0.374}           \\

\textbf{Few-Shot}  & Gemma-2 (9B)    & 0.705&  0.390&  0.568&  0.593&  0.572&  0.571&  0.586&  0.578&  0.526  \\

& Gemma-2 (27B)    &  0.646&  0.231 & 0.670&  0.572&  0.570&  0.570&  0.585 & 0.569&  0.511 \\

& Qwen-2.5 (14B)    &  0.699&  0.257&  \textbf{0.741}&  \underline{0.676}&  0.646 & 0.680&  \textbf{0.656}&  \textbf{0.638}&  \textbf{0.608} \\

& Qwen-2.5 (72B)    & 0.721&  0.400&  \underline{0.736}&  0.503 & 0.617&  0.635&  0.608&  0.626&  0.592 \\

& GPT-4                                 & \multicolumn{1}{c}{\textbf{0.794}}          & \multicolumn{1}{c}{\textbf{0.520}}           & \multicolumn{1}{c}{{0.728}}             & \multicolumn{1}{c}{{0.653}}            &   \textbf{0.680}     &   
\multicolumn{1}{c}{\textbf{0.745}} & \multicolumn{1}{c}{{0.492}} & \multicolumn{1}{c}{{0.473}} & \multicolumn{1}{c}{{0.559}} \\ 

& GPT-4o   &   \underline{0.783}& \underline{0.507}& 0.683& 0.641& \underline{0.664}& 0.730& 0.559& 0.449& 0.529\\

& GPT-4o-mini   &  0.763& 0.435& 0.705& \textbf{0.700}& 0.657& \underline{0.741}& 0.430& 0.404& 0.588 \\

\midrule


& LLaMA-3 (8B)                         & \multicolumn{1}{c}{0.935}          & \multicolumn{1}{c}{0.901}          & \multicolumn{1}{c}{0.935}          & \multicolumn{1}{c}{0.928}          & 0.926         
& \multicolumn{1}{c}{0.935}          & \multicolumn{1}{c}{0.820}          & \multicolumn{1}{c}{0.930}          & \multicolumn{1}{c}{0.924}                   \\ 

& LLaMA-3.1 (8B)& \textbf{0.946} &	0.919&	\underline{0.944}&	\underline{0.934}&	\underline{0.941}&	\underline{0.953}&	\underline{0.850}&	\textbf{0.939}&	\textbf{0.945} \\   

\textbf{Fine-Tuing} & Mistral-v0.3 (7B) & \underline{0.944}&	\underline{0.921}&	\textbf{0.947}&	\textbf{0.935}&	\textbf{0.942}&	\textbf{0.956}&	\textbf{0.852}&	\underline{0.937}&	\underline{0.941} \\  

& Vicuna (7B)                           & \multicolumn{1}{c}{0.937}          & \multicolumn{1}{c}{0.907}          & \multicolumn{1}{c}{0.940}          & \multicolumn{1}{c}{0.906}          & 0.932       
& \multicolumn{1}{c}{\textbf{0.956}} & \multicolumn{1}{c}{{0.823}} & \multicolumn{1}{c}{\textbf{0.936}} & \multicolumn{1}{c}{{0.939}} \\

& Vicuna (13B)                          & \multicolumn{1}{c}{{0.942}} & \multicolumn{1}{c}{\textbf{0.923}} & \multicolumn{1}{c}{0.939}          & \multicolumn{1}{c}{0.923}          & \textbf{0.933} 
& \multicolumn{1}{c}{0.950}          & \multicolumn{1}{c}{{0.847}}          & \multicolumn{1}{c}{{0.935}} & \multicolumn{1}{c}{{0.940}}           \\

\midrule
\midrule
\multicolumn{11}{c}{\textit{Existing Attribution Evaluators}} \\

& \textsc{AutoIS (11B)}                       & \multicolumn{1}{c}{0.609}          & \multicolumn{1}{c}{-}             & \multicolumn{1}{c}{-}             & \multicolumn{1}{c}{-}             & -        & \multicolumn{1}{c}{-}          & \multicolumn{1}{c}{-}          & \multicolumn{1}{c}{-}          & \multicolumn{1}{c}{-}                  \\ 

&  \textsc{AttrScore (13B)}                    & \multicolumn{1}{c}{0.687}          & \multicolumn{1}{c}{-}             & \multicolumn{1}{c}{0.523}          & \multicolumn{1}{c}{0.541}           & 0.521         
 & \multicolumn{1}{c}{0.559}          & \multicolumn{1}{c}{0.410}          & \multicolumn{1}{c}{0.432}          & \multicolumn{1}{c}{0.353}        \\ 
 
\bottomrule
\end{tabular}
}
\caption{Overall results on CAQA. Bold and underlined values are the best and second-best performance, respectively. “-” indicates that the method is not applicable. More complete results can be found in Table \ref{tab:fullresults}. 
}
\label{tab:overallresults}
  \end{minipage}
\end{table*}

Table \ref{tab:overallresults} shows the results of the attribution evaluators on CAQA. Appendix \ref{appendix:CompleteResults} shows more experiments. Our analysis is as follows:

\noindent\textbf{All the evaluators perform poorly in identifying nuanced negative attribution categories, especially \textit{partially supportive}, 
under the zero-shot and few-shot settings. 
} 
Smaller LLMs ($\leq$ 14B), except for Gemma-2 and Qwen-2.5, perform extremely poorly on all three negative categories, suggesting that none of them are capable of distinguishing subtle differences between negative attributions.
Larger LLMs ($\geq$ 27B) demonstrate strong performance on certain negative categories but consistently underperform on \textit{partially supportive} category. For example, in both zero-shot and few-shot settings, the highest scores—0.456 and 0.52, respectively—are achieved by GPT-4. This highlights the challenges faced by the attribution evaluator in accurately identifying \textit{partially supportive} category. 
We find that evaluators often classify \textit{partially supportive} as \textit{supportive}, even though it is apparent that part of the information is missing.

\noindent\textbf{Fine-tuning effectively enhances evaluator performance, while few-shot prompting is particularly beneficial for large-scale evaluators.}
All fine-tuned evaluators trained on our dataset achieve similar performance, exceeding 90\%, except for Mistral-v0.2. AutoIS and AttrScore, which were fine-tuned on other benchmarks, demonstrate limited ability to identify fine-grained attribution categories. Large-scale LLMs ($\geq$ 70B) and GPT-series models exhibit significantly stronger in-context learning capabilities. Under few-shot settings, these models improve by an average of 4.84\% in distinguishing subtle attribution differences, demonstrating deeper contextual understanding. However, few-shot prompting provides only marginal performance gains for smaller LLMs and can even be detrimental to certain models.
See more evaluation of the few-shot setting in Appendix \ref{appendix:CompleteResults}.

\label{cooccurrenceFinding}
\noindent\textbf{The evaluators are often impacted by the keyword co-occurrence and fail to capture the reasoning in  complex attribution scenarios.}
Considering examples of \textit{irrelevant}: the question is “What is the soundtrack of the video game X?” The answer is, “The video game X's soundtrack is Y,” and the evidence is, “Z is a video game designer who has designed games such as X”. The evaluators often incorrectly treats the attribution as \textit{supportive} due to the co-occurring keywords “video game” and “X”, neglecting the semantics of the relation “Soundtrack\_Of”. 
This issue is more serious in the \textit{partially supportive} category, where many instances are incorrectly identified as \textit{supportive}. This misclassification reduces the true positives for \textit{partially supportive} cases while increasing false positives for \textit{supportive} ones. Moreover, apart from GPT models, most other evaluators struggle to understand reasoning logic  that requires integrating multiple pieces of evidence to prove the answer.
The observation can be verified by the fact that all the evaluators usually perform worse on complex attribution with \textit{concatenation}, \textit{intersection} and \textit{union} and on less complex attribution.

\vspace{-0.15cm}
\subsection{Consistency with Human Annotations}\label{sec:cbh}

\begin{figure}
  \centering
  \includegraphics[width=0.48\textwidth]{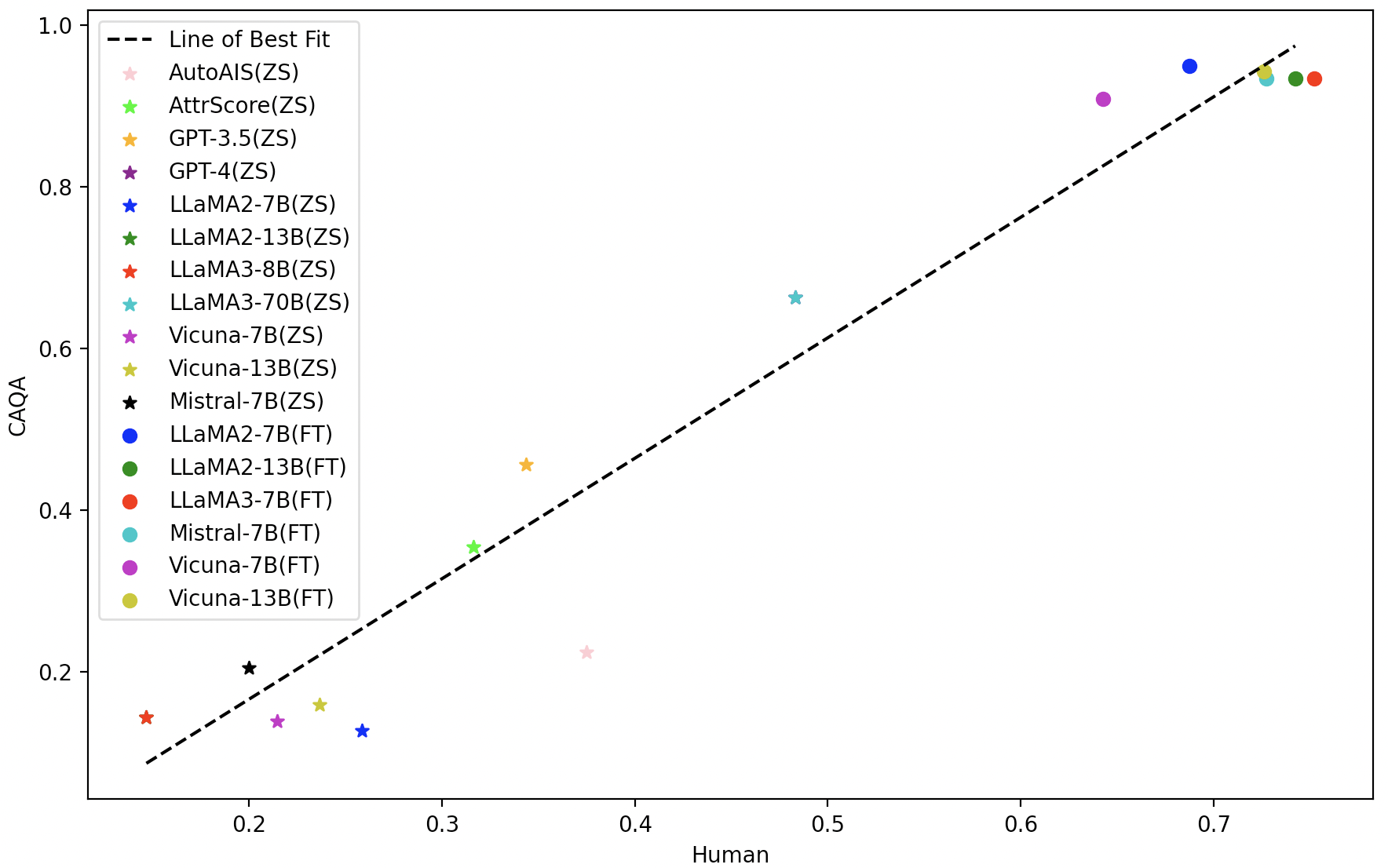}
  \caption{
  Micro-F1 scores of the evaluators on CAQA based on (1) CAQA's own categories (y-axis) and (2) human-annotated categories (x-axis). 
  } 
  \label{fig:ComparisonWithHuman}

\end{figure}



We assess the consistency between the automatically generated categories in CAQA and new categories annotated by humans. With these two sets of categories as gold standard, we compute and compare the overall micro-F1 scores of each evaluator. Some of the results as shown in Figure \ref{fig:ComparisonWithHuman}.
The results demonstrate that the evaluators mostly have consistent results between the two kinds of categories, with a Pearson correlation coefficient of 0.97.
This confirms that evaluation using our automatically generated categories closely align with manual evaluation. 

\vspace{-0.15cm}
\subsection{Fine-grained Evaluation on Partially Supportive}\label{sec:EvalonPS}

Our CAQA benchmark enables a more fine-grained evaluation of attribution, particularly for \texttt{partially supportive} cases, compared to existing benchmarks. To achieve this, we employ the metric \textsc{FactScores} \citep{emnlp/MinKLLYKIZH23}, which is the proportion of sub-facts in the answer that are supported by the citations.
For each \texttt{partially supportive} case, human annotators identify the supported sub-facts, allowing us to calculate \textsc{FactScores} manually. Additionally, CAQA can automatically compute \textsc{FactScores} by comparing the triples in the initial subgraph with those in the subgraph after deletion.
Various automatic evaluators, including LLM evaluators, as well as existing methods such as AutoIS and AttrScore, generate their \textsc{FactScores} by first converting the triples in the initial subgraph $\mathcal{G}$ into natural language sub-facts using ChatGPT. Then, we apply the Retrieve$\rightarrow$LM method \citep{emnlp/MinKLLYKIZH23} to obtain \textsc{FactScores} (see details in  Appendix \ref{appendix:Finegrained_Eval_on_Partially_Supportive}).

\begin{table}
\centering
\small
\caption{\textsc{FactScores} results on 200 partially supportive samples in CAQA. Hum-Gap measures the discrepancy of \textsc{FactScores} 
 w.r.t. human annotations (i.e., 0.58). Fine-tuning uses the training set of CAQA. 
}
    \label{tab:Factscore}
\scalebox{0.78}{
\begin{tabular}{ll|cc}
\toprule
& \multicolumn{1}{l|}{{\textbf{Evaluators}}} & \textsc{FactScores} & Hum-Gap   \\ 
\midrule

 & LLaMA-3 (70B)        & 0.85     &   0.27     \\
 \textbf{Zero-Shot} & GPT-3.5-turbo        & 0.93  & 0.35          \\
 & GPT-4        & 0.84         & 0.26   \\
 \midrule
 & LLaMA-3 (8B)        & 0.19    & 0.39         \\
\textbf{Fine-Tuning} & Vicuna (7B)        & 0.19     & 0.39         \\
 & Vicuna (13B)        & 0.18    & 0.40         \\
 \midrule
 & \textsc{AutoIS} (11B)        &    0.44    &  0.14   \\
 & \textsc{AttrScore} (13B)       &   0.25   &    0.33     \\
 \midrule
 \midrule
 & CAQA Annotations         & 0.62     & \textbf{0.04}        \\
 & Human Annotations        & 0.58    & -         \\

    \bottomrule
    \end{tabular}
    }
\end{table}

The results in Table \ref{tab:Factscore} reveal a significant performance gap between the selected evaluators and human annotations.
The three evaluators fine-tuned on CAQA, which support four categories, and AttrScore, which identifies three, exhibit worse performance (much higher Hum-Gap) compared to AutoIS which identifies only two categories.
Meanwhile, the evaluators in the zero-shot setting have much higher \textsc{FActScores} than human annotations, which reflects the finding in  Section \ref{cooccurrenceFinding}, i.e., their attribution is impacted by  keyword co-occurrence in sub-facts and citations. Additionally, the \textsc{FActScores} of the automated annotations in CAQA differ from human annotations by only 4\%, demonstrating that CAQA is a reliable benchmark for automated fine-grained evaluation.

\vspace{-0.15cm}
\subsection{Exploration of Out-of-Domain Data}
\label{sec:ExplorationOfOOD}
\vspace{-0.05cm}

\begin{table}
\centering
\small
\caption{Performance on ALCE-FineGrained by (1) T5-11B* and Vicuna-13B* and (2) AutoIS and AttrScore. }
\label{table7}
\scalebox{0.75}{
\begin{tabular}{l|ccccc}
\toprule
\multicolumn{1}{l|}{\multirow{2}{*}{\textbf{Evaluators}}} & \multicolumn{5}{c}{\textbf{ALCE-FineGrained}}                                                                                                                          \\ \cmidrule{2-6} 
\multicolumn{1}{c|}{}                                  & \multicolumn{1}{c}{Sup.}       & \multicolumn{3}{c}{Non-Sup.}      & Overall       \\ 
\midrule

AutoIS (T5-11B)                                        & \multicolumn{1}{c}{{0.31}}          & \multicolumn{3}{c}{0.65}                & 0.54         \\ 

T5-11B*                                         & \multicolumn{1}{c}{\textbf{0.44}}          & \multicolumn{3}{c}{\textbf{0.72}}                & \textbf{0.63}          \\ 

\midrule
\midrule
& \multicolumn{1}{c}{Sup.}       & \multicolumn{1}{c}{Par.}    & \multicolumn{1}{c}{Con.} & \multicolumn{1}{c}{Irr.}  & Overall \\
\cmidrule{2-6}

AttrScore (Vicuna-13B)                                               & \multicolumn{1}{c}{{0.52}} & \multicolumn{1}{c}{-}             & \multicolumn{1}{c}{{0.21}} & \multicolumn{1}{c}{{0.42}} & {0.36}         \\

Vicuna-13B*                                 & \multicolumn{1}{c}{{0.54}} & \multicolumn{1}{c}{{0.24}} & \multicolumn{1}{c}{{0.30}} & \multicolumn{1}{c}{{0.34}}          & {0.38} \\

Vicuna-13B* Fine-Tuning                                    & \multicolumn{1}{c}{{\textbf{0.69}}} & \multicolumn{1}{c}{\textbf{0.36}} & \multicolumn{1}{c}{\textbf{0.40}} & \multicolumn{1}{c}{\textbf{0.46}}          & \textbf{0.52} \\ 

\bottomrule
\end{tabular}
}
\vspace{-0.2cm}
\end{table}

We compare the existing evaluators AutoIS (based on T5-11B) and AttrScore (based on Vicuna-13B) that are trained by some other benchmarks, with  T5-11B and Vicuna-13B fine-tuned by CAQA (denoted as T5-11B* and Vicuna-13B*), by testing them on OOD benchmark ALCE-FineGrained.
We also further fine-tune Vicuna-13B* with a small number of samples in ALCE-FineGrained.
The results are in Table \ref{table7}.
Note for comparison with AutoIS, we merge the three negative categories into Non-Supportive.
Compared to AutoIS and AttrScore,  T5-11B* and Vicuna-13B*, have competitive performance in individual classes and the overall score,
demonstrating the effectiveness of CAQA for developing attribution evaluators. Table \ref{table7} also verifies that fine-tuning with a few samples of the domain of the testing dataset further improve the evaluators. See more details in Appendix \ref{appendix:CaseOnALCE-Manual}.


\vspace{-0.2cm}
\section{Conclusion and Future Work}

This work has advanced the evaluation of AQA by proposing a new benchmark CAQA containing comprehensive attribution categories and different attribution complexity levels, and an automatic and general benchmark construction method.
Extensive evaluation has been conducted over 25 automatic evaluators, including two specifically developed baselines, which has demonstrated the effectiveness of CAQA for evaluator assessment and development, and leads to a series of important findings in AQA, such as the shortage of logical reasoning capabilities in dealing with complex attribution scenario. 
In the future, we will study more robust attribution evaluators with CAQA and its variants from other KGs, supporting logical reasoning.

\section*{Limitations}

Although we consider comprehensive attribution categories and complex reasoning scenarios, some more situations including lengthy answers and citations, mathematical, temporal and spatial reasoning, and reasoning requiring domain specific knowledge could be considered in the future.
%
%
For illustration, consider the question: ``When did England last reach the quarterfinals of the World Cup?'' The provided answer is ``England last made the quarterfinals in 1990,'' with a citation noting that ``The England national football team finished in fourth place in 2018 and reached the semifinals in 1990.'' To accurately attribute the answer, it is essential to understand that finishing in fourth place implies participation in the quarterfinals and that 2018 is more recent than 1990.
All these features could be reflected in new versions of our benchmark via considering corresponding KGs (or ontologies) and queries in our automatic construction method.
\section*{Acknowledgments}
This work is partially supported by National Nature Science Foundation of China under No. 62476058, and by the EPSRC project OntoEm (EP/Y017706/1). We gratefully acknowledge the Big Data Computing Center of Southeast University, the Edinburgh International Data Facility (EIDF), and the Data-Driven Innovation Programme at the University of Edinburgh for providing computational resources and support for the numerical experiments conducted in this paper.


\bibliography{acl_latex}

\appendix

\section{Generation of Natural Language Questions, Answers and Attributions}
\label{appendix:GerationOfQAA}

This section presents examples for generating natural language questions, answers, and citations using the GPT-3.5-turbo model. Our approach involves using distinct prompts for transforming subgraphs into comprehensible natural language citations, extending original questions, and converting answer entities into detailed answer statements. Table \ref{tab:kg_to_text} demonstrates the conversion of knowledge graph subgraphs into natural language citations.
Table \ref{tab:question_extend} illustrates the example of generating the extended question.
Table \ref{tab:Answer_statement} provides an example of how answer entities are transformed into long-form answer statements.

Additionally, we verify the quality of the KG-to-natural language text generated by ChatGPT. We ensure the quality of the natural language text generated by focusing on grammatical coherence and content accuracy. Building on prior work \cite{axelsson-skantze-2023-using}, which demonstrates that ChatGPT excels in KG-to-Text tasks with high grammatical correctness and coherence, we primarily evaluate content accuracy to ensure consistency with the corresponding triples. Specifically, we adopted the evaluation framework outlined in the work \cite{axelsson-skantze-2023-using}, assessing whether the generated text accurately reflects the triples. We randomly sampled 100 examples (half of the examples are from generated natural language citations, and the other half are from generated extended questions) and employed two independent annotators to label each instance according to one of three exclusive categories: (1) \textbf{Full Coverage}: The text fully and correctly states all triples. (2) \textbf{Absent}: The text misses some triples. (3) \textbf{Hallucinated}: The text introduces content that actively contradicts the triples. The results are shown in Table \ref{table:annotation}.

\begin{table}[htbp]
\centering
\small
\begin{tabular}{l|c|c|c}
\hline
\textbf{Annotator} & \textbf{Full Coverage} & \textbf{Absent} & \textbf{Hallucinated} \\ 
\hline
\textbf{A} & 92 & 5 & 3 \\ 
\textbf{B} & 95 & 3 & 2 \\ 
\hline
\end{tabular}
\caption{The human evaluation results of the KG-to-natural language text generated by ChatGPT.}
\label{table:annotation}
\end{table}

The Cohen's Kappa score is 0.758, indicating substantial agreement between two annotators. The annotation results demonstrate that ChatGPT reliably generates accurate and coherent text within our benchmark.

\begin{table}[h]
\small
\begin{tcolorbox}

\textbf{Instruction:} Your task is to convert a specific subgraph from the knowledge graph into one or more coherent sentences that summarize the information encapsulated within the subgraph. \\

\textbf{Subgraph}: [(``Wii/DS NA'', ``type'', ``computer\_game\_region''), (``LostWinds: Winter of the Melodias'', ``type'', ``game\_version''), (``LostWinds: Winter of the Melodias'', ``game\_version.regions'', ``Wii/DS NA''), (``LostWinds: Winter of the Melodias'', ``game\_version.distributed\_through'', ``WiiWare''), (``Frontier Developments'', ``cvg\_developer.game\_versions\_developed'', ``LostWinds: Winter of the Melodias'')]. \\

\textbf{Sentences:} \textcolor{blue}{LostWinds: Winter of the Melodias is a computer game with a specific version tailored for the Wii/DS NA region. This version is distributed through WiiWare. The game was developed by Frontier Developments, showcasing their involvement in creating various game versions.}

\end{tcolorbox}
\vspace{-2mm}
\caption{An example about converting a subgraph to a natural language citation using GPT-3.5-turbo.}
\label{tab:kg_to_text}
\end{table}

\begin{table}[h]
\small
\begin{tcolorbox}

\textbf{Instruction:} 
Given knowledge graph triples, your task is to generate a question using all the triples. The generated questions should contain all the relationships.  
 \\

\textcolor{gray}{\# Extended Triples}

\textbf{Triples:} [(?x, type, cvg.computer\_videogame), (?x, computer\_videogame.influenced\_by, Sengoku Rance), (?x, fictional.setting, Touhou Project)] \\

\textbf{Extended question:} \textcolor{blue}{What computer video game was influenced by Sengoku Rance and is set in the Touhou Project fictional universe?}

\end{tcolorbox}
\vspace{-2mm}
\caption{An example about generating the extended question using GPT-3.5-turbo.}
\label{tab:question_extend}
\end{table}

\begin{table}[h]
\small
\scalebox{1}{%
\begin{tcolorbox}

\textbf{Instruction:} Your task is to convert a question along with its concise answer into a comprehensive answer statement.
\\

\textbf{Question:} What group fought in the Battle of Vicksburg that was based in Montgomery? \\
\textbf{Answer:}  Army of Mississippi \\

\textbf{Answer statement:} \textcolor{blue}{The group that fought in the Battle of Vicksburg and was based in Montgomery was the Army of Mississippi.}

\end{tcolorbox}
}
\vspace{-2mm}
\caption{An example about converting the answer entity to a long answer statement using GPT-3.5-turbo.}
\label{tab:Answer_statement}
\end{table}

\section{Concrete Cases of Query Extension}
\label{appendix:ConcreteCasesofQueryExtension}

We incorporate concrete examples alongside explanations of each query extension strategy described in Section 4.2. These examples correspond to the rules summarized in Table 3 and illustrate the process of each extension strategy.

\textbf{Case 1: Single-triple Query with the Union Extension Strategy}. The grounded graph of the original query $l$: 
(Rick Scott (Q439729), place of birth, Bloomington). By querying the KG for other entities with the same name (e.g., Rick Scott), we may discover an additional entity \texttt{Rick Scott (Q7329369)} and retrieve its place of birth: 
(Rick Scott (Q7329369), place of birth, Stone Mountain). The grounded graph of the extended query $l'$ after Union extension becomes:
[(Rick Scott (Q439729), place of birth, Bloomington), (Rick Scott (Q7329369), place of birth, Stone Mountain)].

\textbf{Case 2: Path-like Query with Intersection Extension Strategies}. The grounded graph of the original query $l$: 
[(Rick Scott, place of birth, Bloomington), (Bloomington, capital of, McLean County)].

\textit{First Intersection Strategy (Head Entity Extension):}
We extend the query by adding a triple related to the head entity Rick Scott, e.g.:
(Anna Scott, spouse, Rick Scott).
Then the grounded graph of the extended query $l'$:
[(Anna Scott, spouse, Rick Scott), (Rick Scott, place of birth, Bloomington), (Bloomington, capital of, McLean County)]. 

\textit{Second Intersection Strategy (Answer Entity Extension):}  
We extend based on the answer entity Bloomington, adding:
(McLean County, country, United States).

Then the grounded graph of the extended query $l'$ becomes:
[(Rick Scott, place of birth, Bloomington), (Bloomington, capital of, McLean County), (McLean County, country, United States)].

\textbf{Case 3: Tree-like Query (Intersection Extension Strategies)}.
The grounded graph of the original query $l$: 
[(Rick Scott, educated at, University of Missouri–Kansas City), (Sharice Davids, educated at, University of Missouri–Kansas City)].

\textit{First Intersection Strategy (Head Entity Extension):} 
Add a triple involving the head entity Rick Scott, such as:
(Anna Scott, spouse, Rick Scott).
Then the grounded graph of the extended query $l'$:
[(Anna Scott, spouse, Rick Scott), (Rick Scott, educated at, University of Missouri–Kansas City), (Sharice Davids, educated at, University of Missouri–Kansas City)].

\textit{Second Intersection Strategy (Answer Entity Extension):}  
Extend the query by adding a triple about the answer entity University of Missouri–Kansas City:
(University of Missouri–Kansas City, located in, Kansas City).
Then the grounded graph of the extended query $l'$:
[(Rick Scott, educated at, University of Missouri–Kansas City), (Sharice Davids, educated at, University of Missouri–Kansas City), (University of Missouri–Kansas City, located in, Kansas City)].

\section{CAQA Benchmark Construction and Statistics}
\label{appendix:CAQAConstruction}

The CAQA benchmark is built on the top of two KGQA datasets, GrailQA\footnote{GrailQA is licensed under CC BY-SA 4.0.} and WebQuestionsSP, with the knowledge graph Freebase, forming a comprehensive attribution evaluation testbed.
We selectively include samples from these two datasets whose logical queries align with single-triple, path-like, or tree-like queries, as delineated in Section 4.1. 
For path queries, we collect the example with a path length of at most two hops. We treat paths incorporating CVT (Compound Value Type) nodes as one-hop. For example, {[\textit{(Harper Lee, person.education $?cvt$), ($?cvt$ education.institution, Monroe County High School)}]} is a one-hop path, where the node $?cvt$ holds no actual meaning. Regarding tree-liked queries, we restrict our selection to those with a maximum of two non-answer nodes, meaning up to two subject entities.

\begin{figure*}[ht]
    \centering
    \begin{minipage}[b]{0.45\linewidth}
        \includegraphics[width=\linewidth]{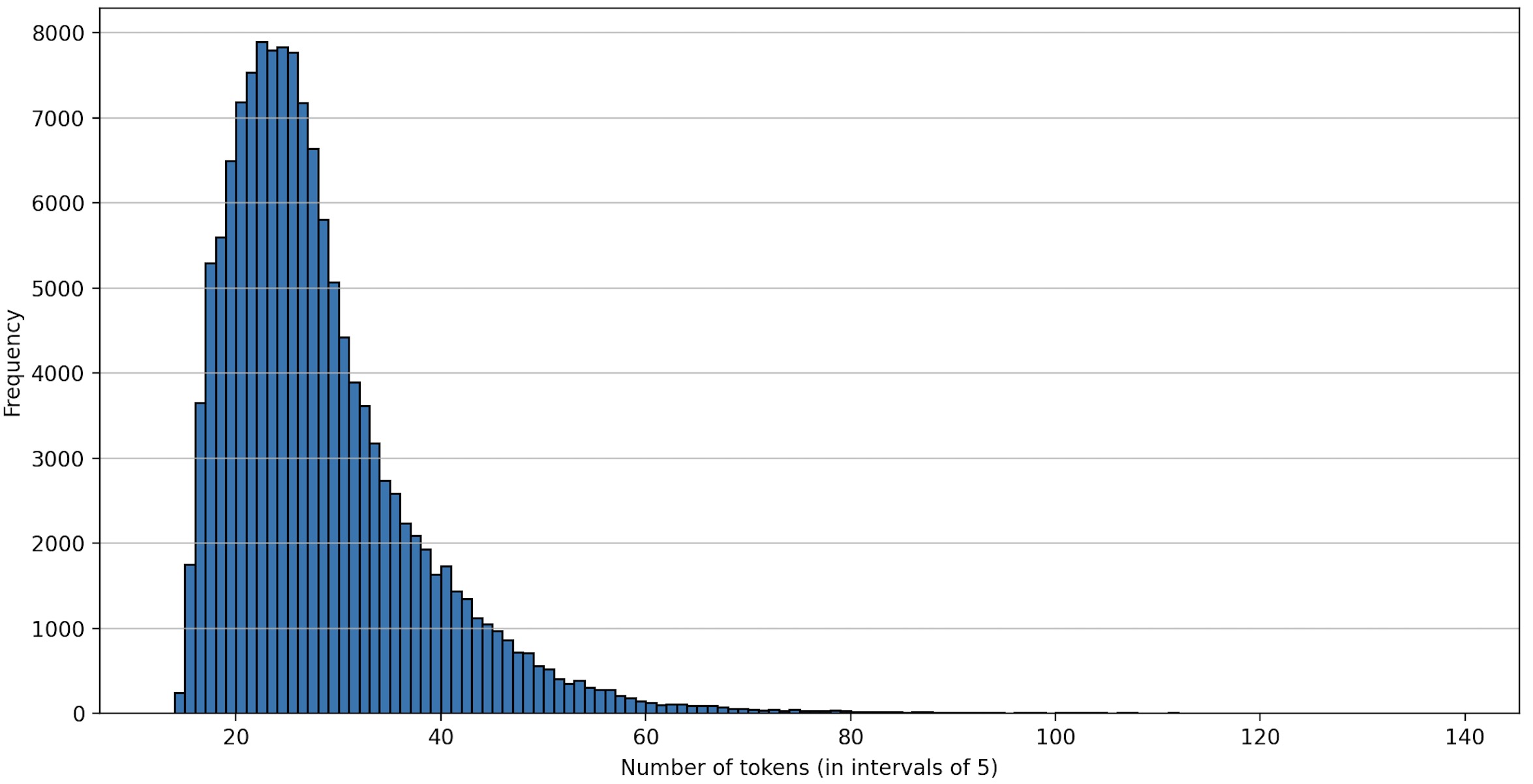}
        \caption{Histogram of the number of tokens across all citations in the CAQA benchmark training set.}
        \label{fig:image1}
    \end{minipage}
    \hfill 
    \begin{minipage}[b]{0.45\linewidth}
         \includegraphics[width=\linewidth]{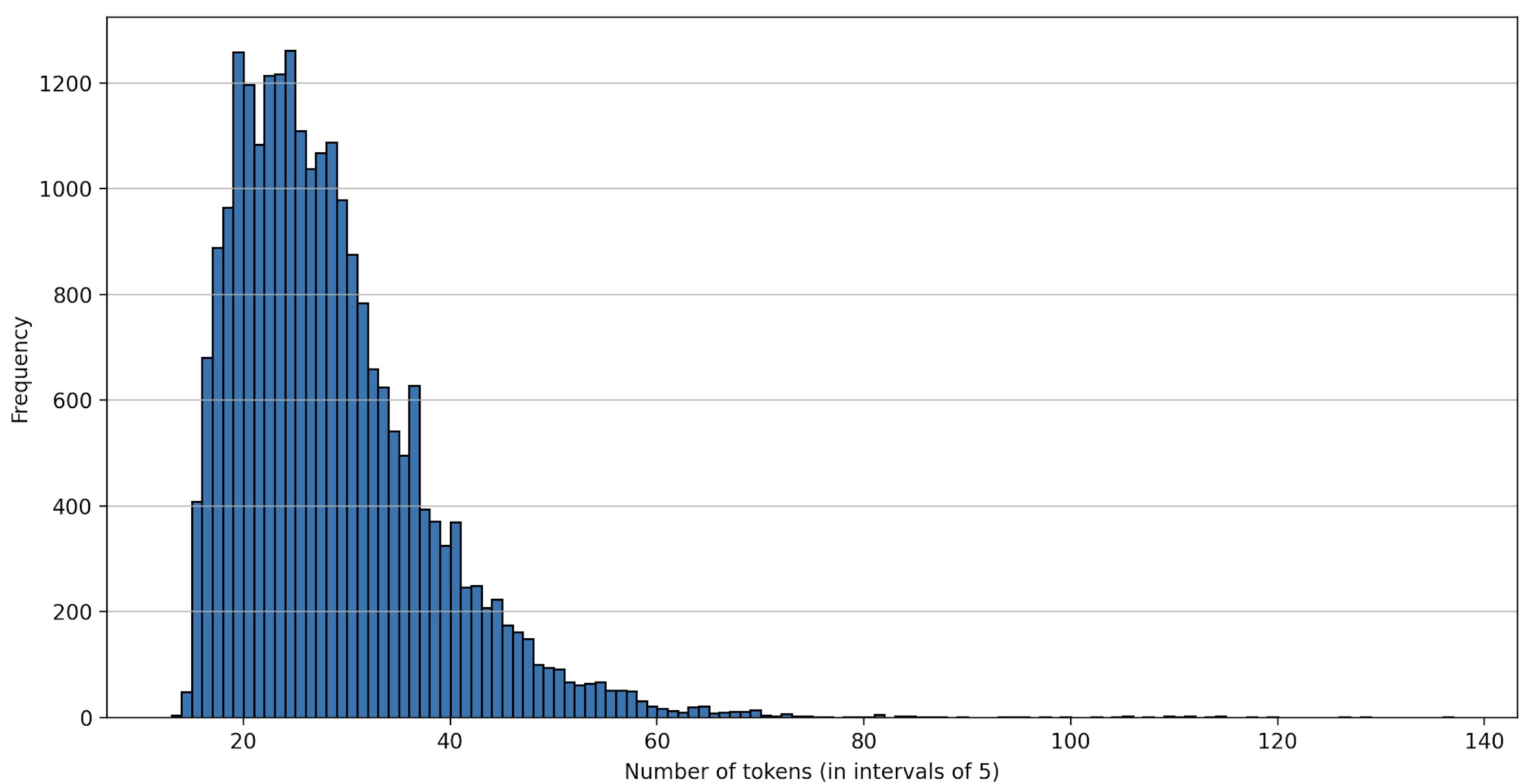} 
        \caption{Histogram of the number of tokens across all citations in the CAQA benchmark test set.}
        \label{fig:image2}
    \end{minipage}
\end{figure*}

\begin{figure*}[h]
  \centering
  \includegraphics[width=.95\textwidth]{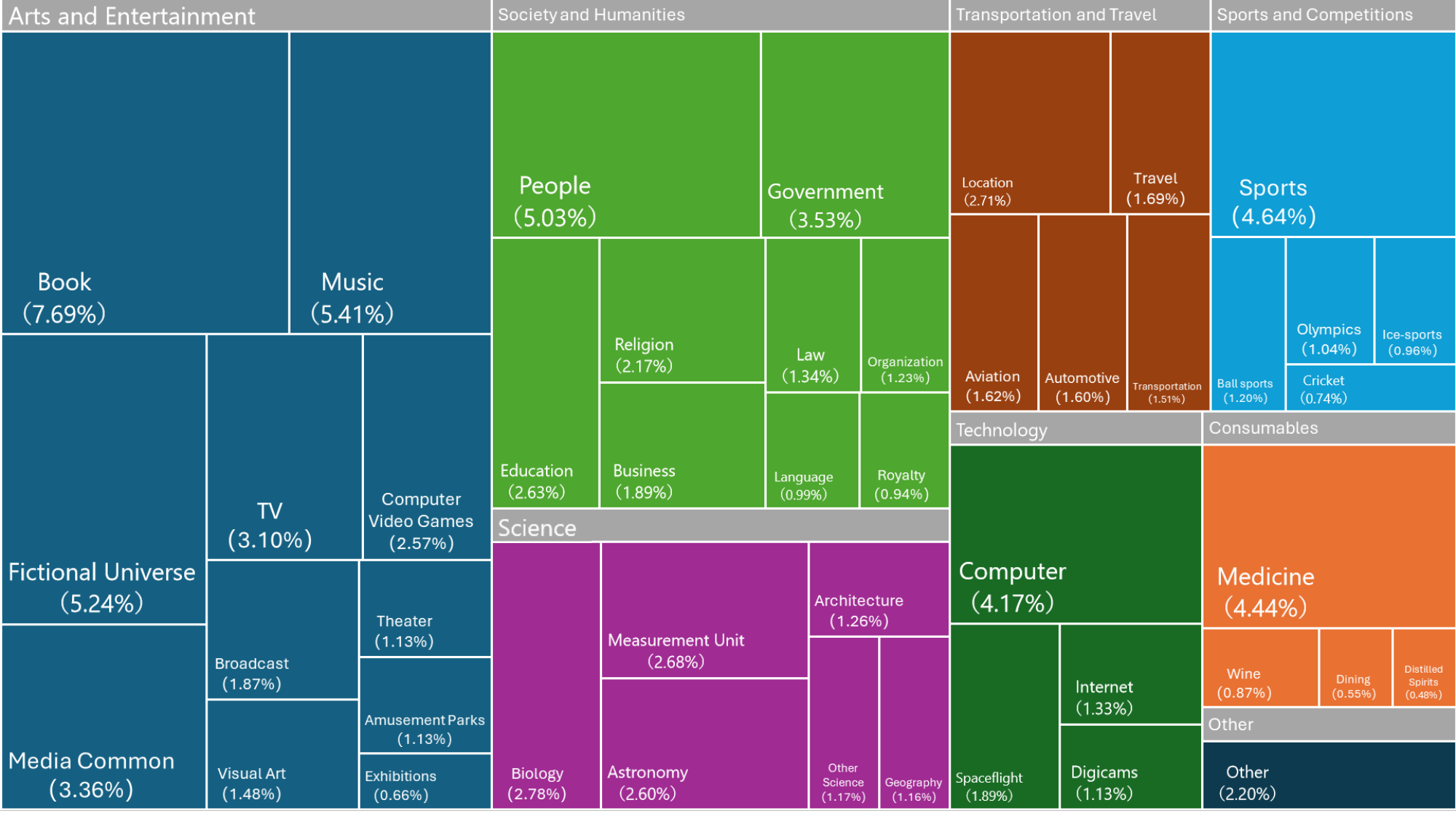}
  \caption{The distribution of examples across different domains in the CAQA benchmark.
  } 
  \label{figure3}
\end{figure*}

The length distribution (i.e., the number of tokens) of citations in the training and test sets of the CAQA benchmark is depicted in Figures \ref{fig:image1} and \ref{fig:image2}. 
These distributions reveal a concentration of citations around 25 tokens, with a minority exceeding 60 tokens.  In future work, we aim to enhance the complexity and length of natural language references by developing more intricate subgraphs. 
Additionally, Figure \ref{figure3} presents the domain distribution within the CAQA benchmark. This distribution underscores the benchmark's broad domain coverage and its encompassment of various sub-domains, highlighting the diversity of our benchmark.

\section{Implementation Details}
\label{appendix:ImplementationDetails}

\subsection{Details of Evaluators and Prompts}

\begin{table*}[h]
\small
\begin{tcolorbox}

\textbf{GPT series}

\textbf{Instruction:} Your task is to evaluate the relationship between a provided citation and the answer to a specific question. There are four possible types of relationships:

1. Supportive: Choose this if the citation directly confirms or is fully in alignment with the answer, providing all necessary information to substantiate it.

2. Insufficient: Choose this when the citation provides only partial backing for the answer, lacking some essential details or evidence needed for full support.

3. Contradictory: Choose this option if the citation is consistent with the intent of the question but directly opposes or contradicts the answer. 

4. Irrelevant: Select this option if the citation does not match the intent of the question and contains information that is not useful for answering. 

For each example provided: First, you need to look at the question given and the answer provided. Then, compare them with the content of the citation. Finally, select the appropriate relationship category based on whether the citation supports the answer, is missing information, contradicts itself, or is irrelevant to the answer.

Example:

\textbf{Question:} \{question\}

\textbf{Answer:} \{answer statement\}

\textbf{Reference:} \{citation\}

\textbf{Relationship Category:}

\rule{\linewidth}{0.2mm}
\textbf{AttrScore}

\textbf{premise:} \{question$|$answer statement\}

\textbf{hypothesis:} \{citation\}

\rule{\linewidth}{0.2mm}

\textbf{AutoIS}

Below is an instruction that describes a task, paired with an input that provides further context. Write a response that appropriately completes the request.
        
\textbf{Instruction:} Verify whether a given reference can support the claim. Options: Attributable, Extrapolatory or Contradictory.

\textbf{Claim:} \{question$|$answer statement\}

\textbf{Reference:} \{citation\}

\textbf{Response:}

\rule{\linewidth}{0.2mm}
\textbf{Other Evaluators}

Below is an instruction that describes a task, paired with an input that provides further context. Write a response that appropriately completes the request.
        
\textbf{Instruction:} Verify whether a given reference can support the claim. Options: Supportive, Insufficient, Contradictory or Irrelevant.

\textbf{Claim:} \{question$|$answer statement\}

\textbf{Reference:} \{citation\}

\textbf{Response:}

\end{tcolorbox}
\vspace{-2mm}
\caption{Different prompts designed for different evaluators.}
\label{tab:prompt}
\end{table*}

\begin{table*}[h]
\small
\begin{tcolorbox}

\textbf{GPT-3.5 series}

\textbf{Instruction:} 
Your task is to evaluate the relationship between a provided citation and the answer to a specific question. There are four possible types of relationships:

1. Supportive: Choose this if the citation directly confirms or is fully in alignment with the answer, providing all necessary information to substantiate it.

2. Insufficient: Choose this when the citation provides only partial backing for the answer, lacking some essential details or evidence needed for full support.

3. Contradictory: Choose this option if the citation is consistent with the intent of the question but directly opposes or contradicts the answer. 

4. Irrelevant: Select this option if the citation does not match the intent of the question and contains information that is not useful for answering. 

Please read the examples and choose the most appropriate relationship category for the test example.

Example 1:
\{Support Example\}

Example 2:
\{Missing Example\}

Example 3:
\{Contradictory Example\}

Example 4:
\{Irrelevant Example\}

Test Example:

\textbf{Question:} \{question\}

\textbf{Answer:} \{answer statement\}

\textbf{Reference:} \{citation\}

\textbf{Relationship Category:}

\rule{\linewidth}{0.2mm}
\textbf{Other Evaluators}

Below is an instruction that describes a task, paired with an input that provides further context. Write a response that appropriately completes the request.
        
\textbf{Instruction:} Verify whether a given reference can support the claim. Options: Supportive, Insufficient, Contradictory or Irrelevant.

\{Support Example\}

\{Missing Example\}

\{Contradictory Example\}

\{Irrelevant Example\}

\textbf{Claim:} \{question$|$answer statement\}

\textbf{Reference:} \{citation\}

\textbf{Response:}

\end{tcolorbox}
\vspace{-2mm}
\caption{Different few-shot prompts designed for different evaluators.}
\label{tab:fewprompt}
\end{table*}

AutoIS is a natural language inference (NLI) model\footnote{https://huggingface.co/google/t5\_xxl\_true\_nli\_mixture} based on T5-11B  that outputs a ``1'' to indicate that the citation supports the answer statement or a ``0'' to indicate a lack of support.
AttrScore is a uniform name for attribution evaluators developed on various LLMs, and we use the best-performing attribution evaluator (Vicuna-13B) on the original work for comparison. 
Since AutoIS can only recognise \textit{supportive} and \textit{non-supportive} attribution categories, we only report its F1 score on \textit{supportive} in Table \ref{tab:overallresults}. In the experiments on the ALCE-FineGrained benchmark, to be able to compare the evaluator trained on our benchmark with AutoIS, we merge the three incorrect categories into the \textit{non-supportive} category, and then compute F1 scores of \textit{supportive} and \textit{non-supportive} as well as overall micro-F1 score. 

Table \ref{tab:prompt} describes the different prompt designs against the various attribution evaluators. 
During our experiments, we have explored providing more detailed explanations for each category to non-GPT models. However, we observed that overly detailed prompts often led to a decline in their performance. To maximize the performance of all models and to select the best attribution evaluator, we opted for prompts that were calibrated to achieve the best results for all models.

In the few-shot setting, we select one sample per attribution category as a demonstration, as shown in Table \ref{tab:fewprompt}. We explore on more few-shot settings in Appendix \ref{appendix:CompleteResults}. For model fine-tuning, we use the prompt of ``Other Evaluators''  depicted in Table \ref{tab:prompt} as input of all models, and the output of the model is one of the four attribution categories proposed. We use two A100 80G GPUs for full parameter fine-tuning and two A100 80G GPU for the inference phase  (four A100 80G GPUs for inference with model parameters greater than 70B). During inference, text generation is conducted with a temperature setting of 0 and vLLM \cite{Kwon2023EfficientMM} is used to acceleration. If LLMs produce an attribution category with an explanation, we extract the predicted label using regular expression techniques. Due to high costs, we evaluate GPT models on only 3000 test samples.

\subsection{Implementations of Various Evaluators for Calculating \textsc{FActScores}}
\label{appendix:Finegrained_Eval_on_Partially_Supportive}

We describe how the various attribution evaluators calculate \textsc{FActScores} in the CAQA benchmark.
For each \textit{partially supportive} case, we first use GPT-3.5 to convert triples into natural language subfacts with the prompt: ``Your task is to convert a triple into natural language statement''.
Then, following the Retrieve$\rightarrow$LM method \citep{emnlp/MinKLLYKIZH23}, we calculate the \textsc{FActScores} for each attribution evaluator. Specifically, the prompt is fed into the evaluator, which predicts True or False to calculate the \textsc{FActScores}. For the zero-shot evaluator, we use the prompt: ``\textit{Judge this fact based on the given context.\textbackslash n\textbackslash n Fact: \{sub-fact\}\textbackslash n Text: \{citation\} \textbackslash n\textbackslash nTrue or False?\textbackslash nOutput:}''. For fine-tuned and existing evaluators, the prompt provided in Table \ref{tab:prompt} is used. Because these evaluators can generate more than two attribution categories, we categorize supportive as True and all other categories as False for calculating the \textsc{FActScores}. Human annotation, as described in Appendix \ref{appendix:HumanAnnotation}, involves annotators determining whether each subfact is supported by its citation. The \textsc{FActScores} is the proportion of predictions classified as True compared to the total number of subfacts evaluated.

\section{Detailed Experimental Results}
\label{appendix:CompleteResults}

\subsection{Full Experimental Results}
\label{appendix:subSecFullResults}

\begin{table}
\centering
\small
\vspace{-0.3cm}
\scalebox{0.75}{
\begin{tabular}{l|ccccc}
\toprule
\multicolumn{1}{l|}{{\multirow{2}{*}{\textbf{N-shot (GPT-3.5-turbo)}}}} & \multicolumn{5}{c}{\textbf{CAQA}}                                                                                                                          \\ 
\cmidrule{2-6}

& \multicolumn{1}{c}{Sup.}       & \multicolumn{1}{c}{Par.}    & \multicolumn{1}{c}{Con.} & \multicolumn{1}{c}{Irr.}  & Overall \\

\midrule

1-shot                                              & \multicolumn{1}{c}{{0.613}} & \multicolumn{1}{c}{0.026}             & \multicolumn{1}{c}{{0.318}} & \multicolumn{1}{c}{\textbf{0.609}} & {0.476}         \\

2-shot                                   & \multicolumn{1}{c}{\textbf{0.627}} & \multicolumn{1}{c}{\textbf{0.034}} & \multicolumn{1}{c}{{0.359}} & \multicolumn{1}{c}{{0.593}}          & \textbf{0.486} \\ 
3-shot                                & \multicolumn{1}{c}{{0.599}} & \multicolumn{1}{c}{{0.015}} & \multicolumn{1}{c}{\textbf{0.378}} & \multicolumn{1}{c}{{0.581}}          & {0.478} \\

\bottomrule
\end{tabular}
}
\caption{The performance of GPT-3.5-turbo under various few-shot settings on CAQA.}
\label{tab:Nshot}
\vspace{-0.3cm}
\end{table}

We present the full experimental results in Tables \ref{tab:fullresults}. 
We find that the newest LLMs demonstrate slight improvements in overall performance for attribution recognition compared to their earlier versions in all three settings. Specifically, in zero-shot and few-shot settings, they show more balanced and enhanced results across scenarios with varying levels of complexity, highlighting advancements in their reasoning capabilities. However, they still struggle to accurately distinguish fine-grained attribute categories, with notable deficiencies persisting in certain attribution recognition.

\subsection{Exploration of Few-shot Settings}
\label{appendix:subSecFewshotExploration}
We investigate three few-shot settings: 1-shot, 2-shot, and 3-shot in 5,000 test instances employing GPT-3.5-turbo. In these settings, 1, 2, and 3 examples, respectively, are provided for each attribution category. The outcomes, as displayed in Table \ref{tab:Nshot}, suggest that increasing the number of examples yields negligible improvement in performance. Consequently, considering the associated costs, we have opted to use the 1-shot setting in all subsequent experiments.

\subsection{Exploration of Advanced Baselines}
\label{appendix:subSecAdvancedBaselines}
We have included more advanced and diverse baselines, particularly incorporating LLM-based evaluators with CoT strategies. Specifically, inspired by the approach proposed in \citep{emnlp/MinKLLYKIZH23}, we designed a multi-step CoT evaluation strategy that enhances the model's attribution evaluation capabilities: 

1. Decomposition: The LLM is first prompted to break down the answer into a series of atomic facts. The prompt is: “\textit{Break the following sentence into independent sub-facts. If the sentence is already a minimal fact, return it directly. Each sub-fact should be concise, self-contained, and listed on a new line. \textbackslash n
Sentence: {Sentence}\textbackslash n
Sub-facts:}”.

2. Atomic Fact Attribution Evaluation: For each atomic fact, the LLM assesses its relationship with the evidence, categorizing it as supportive, contradictory, or irrelevant. The prompt is: “\textit{Evaluate the relationship between the statement and reference. Choose one of the following categories based on how the citation relates to the statement:\textbackslash n1. Supportive: Select this if the statement is present and validated by the reference.\textbackslash n
2. Contradictory: Select this if the reference contains the fact that directly opposes or contradicts the statement.\textbackslash n
3. Irrelevant: Select this if the reference is completely unrelated to the statement.\textbackslash n
For each example provided: First, you need to understand precisely what the statement claims. Then, understand what facts or information the reference actually provides. Finally, compare the statement with the information of the reference and select the appropriate relationship category.\textbackslash n\textbackslash n Example:\textbackslash n \textbackslash n Statement: {sub\_fact}\textbackslash n Reference: {reference}\textbackslash n Relationship Category:}”.

3. Final Attribution Classification: Based on the attribution types of the atomic facts, we determine the overall attribution label for the answer as follows:
Supportive: All atomic facts are supported by the evidence.
Partially Supportive: At least one atomic fact is supported, and none are contradictory.
Contradictory: At least one atomic fact is contradicted by the evidence.
Irrelevant: All atomic facts are irrelevant to the evidence.

The experiment results are shown in table \ref{tab:AdvancedBaselines}.
Compared with the naive LLM evaluators (see Table \ref{tab:overallresults}), the experimental results of our proposed CoT-based evaluators reveal two key findings:
1. Effectiveness in Complex Scenarios: The CoT-based method demonstrates strong performance in complex attribution scenarios, such as Concatenation, Intersection, and Union, achieving up to a 16\% improvement over the naive LLM evaluator in complex attribution scenarios. However, in the simpler Single scenario, decomposition errors will hinder performance, resulting in an overall score that remains largely unchanged in the zero-shot setting.
2. Improvement with Few-shot Learning: Under the few-shot setting, the CoT-based evaluator benefits significantly from in-context examples, which help guide both fact decomposition and attribution classification. Compared to the equally naive LLM evaluator, the overall performance of the CoT-based evaluator improved by as much as 5.2\%, highlighting the value of in-context learning in improving the evaluation of attribution.

\begin{table*}[htbp]
  \centering
  \small
  \begin{minipage}[b]{1\textwidth} 
    \centering
\scalebox{0.8}{
\begin{tabular}{llccccc|cccc}
\toprule
\multirow{2}{*}{\textbf{Settings}} & \multirow{2}{*}{\textbf{Evaluators (Size)}} & \multicolumn{5}{c}{\textbf{Category}}  & \multicolumn{4}{|c}{\textbf{Complexity}}    \\ 
\cmidrule{3-7}
\cmidrule{8-11}

& & \multicolumn{1}{c}{Sup.}       & \multicolumn{1}{c}{Par.}       & \multicolumn{1}{c}{Con.} & \multicolumn{1}{c}{Irr.}    & Overall      
 & \multicolumn{1}{c}{S.}       & \multicolumn{1}{c}{C.}     & \multicolumn{1}{c}{I.}   & \multicolumn{1}{c}{U.}      \\ 
\midrule


& Gemma-2 (27B)    & 0.663  & 0.216  & 0.55& 0.597& 0.556&0.557& 0.667& 0.542& 0.375         \\

& LLaMA-3.1 (70B)     & 0.695& 0.166& 0.54& 0.58& 0.537&0.527& 0.681& 0.542& 0.5         \\

\textbf{Zero-Shot} &   Qwen-2.5 (72B)    & 0.653& 0.294& \textbf{0.75}& 0.496& 0.604& 0.585& \textbf{0.744}& \textbf{0.639}& 0.538 \\

& GPT-4o   &    \textbf{0.746}& 0.387& 0.659& 0.649& \textbf{0.624}& \textbf{0.654}& 0.604& 0.528& \textbf{0.688} \\

& GPT-4o-mini  & 0.698& \textbf{0.413}& 0.593& \textbf{0.717}& 0.607& 0.642& 0.5& 0.517&\textbf{0.688} \\
 
\midrule


& Gemma-2 (27B)   & 0.645& 0.293& 0.692& 0.554& 0.581& 0.573& 0.708& 0.592& 0.438        \\ 

& LLaMA-3.1 (70B)     & 0.722&  0.361&  0.714&  0.637&  0.64& 
 0.64&  0.688&  0.637&  0.562         \\ 

\textbf{Few-Shot} &   Qwen-2.5 (72B)    & 0.755& 0.449& 0.771& 0.505& 0.646& 0.636& \textbf{0.708} & \textbf{0.682} & 0.5 \\

& GPT-4o   &   \textbf{0.817} & \textbf{0.615} & 0.719& 0.679& \textbf{0.716} & \textbf{0.765} & 0.542& 0.617& \textbf{0.688} \\

& GPT-4o-mini   &  0.770& 0.497& \textbf{0.786} & \textbf{0.706} & 0.697& 0.741& 0.417& 0.398& \textbf{0.688} \\

\bottomrule
\end{tabular}
}
\caption{Advanced baseline results on CAQA. The bold and underlined results show the best.}
\label{tab:AdvancedBaselines}
  \end{minipage}
\end{table*}

\begin{table*}[htbp]
  \centering
  \small
  \begin{minipage}[b]{1\textwidth} 
    \centering
\scalebox{0.8}{
\begin{tabular}{llccccc|cccc}
\toprule
\multirow{2}{*}{\textbf{Settings}} & \multirow{2}{*}{\textbf{Evaluators (Size)}} & \multicolumn{5}{c}{\textbf{Category}}  & \multicolumn{4}{|c}{\textbf{Complexity}}    \\ 
\cmidrule{3-7}
\cmidrule{8-11}

& & \multicolumn{1}{c}{Sup.}       & \multicolumn{1}{c}{Par.}       & \multicolumn{1}{c}{Con.} & \multicolumn{1}{c}{Irr.}    & Overall      
 & \multicolumn{1}{c}{S.}       & \multicolumn{1}{c}{C.}     & \multicolumn{1}{c}{I.}   & \multicolumn{1}{c}{U.}      \\ 
\midrule

& LLaMA-2 (7B)                          & \multicolumn{1}{c}{0.423}          & \multicolumn{1}{c}{0.121}          & \multicolumn{1}{c}{0.057}           & \multicolumn{1}{c}{0.170}          & 0.279  
& \multicolumn{1}{c}{0.286}          & \multicolumn{1}{c}{0.249}          & \multicolumn{1}{c}{0.282}          & \multicolumn{1}{c}{0.260} \\  

& LLaMA-2 (13B)                         & \multicolumn{1}{c}{0.418}          & \multicolumn{1}{c}{0.164}          & \multicolumn{1}{c}{0.161}          & \multicolumn{1}{c}{0.125}          & 0.279         
& \multicolumn{1}{c}{0.314}          & \multicolumn{1}{c}{0.270}          & \multicolumn{1}{c}{0.303}          & \multicolumn{1}{c}{0.253} \\ 

& LLaMA-3 (8B)                          & \multicolumn{1}{c}{0.467}          & \multicolumn{1}{c}{0.120}           & \multicolumn{1}{c}{0.072}             & \multicolumn{1}{c}{0.007}             & 0.296     & \multicolumn{1}{c}{0.304}          & \multicolumn{1}{c}{0.271}          & \multicolumn{1}{c}{0.283}          & \multicolumn{1}{c}{0.259}     \\ 

& LLaMA-3.1 (8B)                  &  0.544&	0.049&	0.130&	0.017&	0.318&	0.319&	0.326&	0.319&	0.285     \\ 

& LLaMA-3 (70B)                        & \multicolumn{1}{c}{0.746}          & \multicolumn{1}{c}{0.104}           & \multicolumn{1}{c}{0.653}             & \multicolumn{1}{c}{{0.592}}             & 0.525  & \multicolumn{1}{c}{0.645}          & \multicolumn{1}{c}{0.279}          & \multicolumn{1}{c}{0.305}          & \multicolumn{1}{c}{0.578} 
\\ 

& LLaMA-3.1 (70B)  & 0.688& 0.168& 0.547& 0.609& 0.544& 0.545& 0.549& 0.545& 0.499             \\ 

& Mistral-v0.2 (7B)                          & \multicolumn{1}{c}{0.456}          & \multicolumn{1}{c}{0.178}          & \multicolumn{1}{c}{0.191}          & \multicolumn{1}{c}{0.153}          & 0.305        & \multicolumn{1}{c}{0.315}          & \multicolumn{1}{c}{0.281}          & \multicolumn{1}{c}{0.294}          & \multicolumn{1}{c}{0.265}   \\

& Mistral-v0.3 (7B)    & 0.661& 0.160& 0.051& 0.334& 0.362& 0.363& 0.374& 0.356& 0.337   \\ 

& Ministral* (8B)    &- &-&-&-&-&-&-&-&-  \\

& Mixtral-v1.0 (8x7B)    &  0.677& 0.094& 0.17& 0.635& 0.494& 0.495& 0.516& 0.484& 0.487 \\

\textbf{Zero-Shot}  & Vicuna (7B)                           & \multicolumn{1}{c}{0.513}          & \multicolumn{1}{c}{0.100}           & \multicolumn{1}{c}{0.064}           & \multicolumn{1}{c}{0.199}           &   0.327       & \multicolumn{1}{c}{0.343}          & \multicolumn{1}{c}{0.273}          & \multicolumn{1}{c}{0.312}          & \multicolumn{1}{c}{0.256}    \\ 

&  Vicuna (13B)                          & \multicolumn{1}{c}{0.634}          & \multicolumn{1}{c}{0.211}           & \multicolumn{1}{c}{0.393}             & \multicolumn{1}{c}{0.275}             & 0.405  
 & \multicolumn{1}{c}{0.432}          & \multicolumn{1}{c}{0.314}          & \multicolumn{1}{c}{0.361}          & \multicolumn{1}{c}{0.374}  \\ 
 
& Phi-3-small (7B)    & 0.624& 0.217& 0.481& 0.569& 0.507& 0.509& 0.508& 0.504& 0.453  \\

& Phi-3-medium (14B)    &  0.627& 0.148& 0.383& 0.291& 0.406& 0.406& 0.413& 0.407& 0.391 \\

& Gemma-2 (9B)    &   0.667& 0.28& 0.498& 0.624& 0.556& 0.557& 0.572& 0.552& 0.508\\

& Gemma-2 (27B)    &  0.653& 0.184& 0.569& 0.646& 0.566& 0.566& 0.579& \underline{0.566}& 0.537 \\

& Qwen-2.5 (7B)    &  0.696& 0.241& 0.617& 0.404& 0.556& 0.557& 0.572& 0.551& 0.537 \\

& Qwen-2.5 (14B)    &  0.68& 0.132& \underline{0.708}& \underline{0.66}& \underline{0.617}& 0.640& \textbf{0.622}& \textbf{0.611}& 0.547 \\

& Qwen-2.5 (72B)    &  0.629& 0.266& 0.701& 0.471& 0.571& 0.593& 0.583& 0.565& 0.53 \\

& {GPT-3.5-turbo}                               & \multicolumn{1}{c}{0.583}          & \multicolumn{1}{c}{0.017}           & \multicolumn{1}{c}{0.598}          & \multicolumn{1}{c}{0.512}          & 0.497     
& \multicolumn{1}{c}{0.555}          & \multicolumn{1}{c}{0.321}          & \multicolumn{1}{c}{0.363}          & \multicolumn{1}{c}{{0.363}}    \\ 

& GPT-4                                 & \multicolumn{1}{c}{\textbf{0.771}} & \multicolumn{1}{c}{\textbf{0.456}} & \multicolumn{1}{c}{\textbf{0.745}} & \multicolumn{1}{c}{{{0.473}}} & \textbf{0.630} 
& \multicolumn{1}{c}{\textbf{0.685}} & \multicolumn{1}{c}{{0.451}} & \multicolumn{1}{c}{{0.514}} & \multicolumn{1}{c}{\textbf{0.616}} \\ 

& GPT-4o   &  \underline{0.769}& \underline{0.445}& 0.598& 0.626& \textbf{0.630}& \underline{0.676}& \underline{0.591}& 0.47& \underline{0.588} \\

& GPT-4o-mini   & 0.718& 0.297& 0.632& \textbf{0.703}& {0.616}& 0.672& 0.473& 0.444& 0.559 \\

\midrule


& LLaMA-2 (7B)                     & \multicolumn{1}{c}{0.300}          & \multicolumn{1}{c}{0.066}          & \multicolumn{1}{c}{0.009}           & \multicolumn{1}{c}{0.334}           & 0.248          
 & \multicolumn{1}{c}{0.259}          & \multicolumn{1}{c}{0.218}          & \multicolumn{1}{c}{0.167}          & \multicolumn{1}{c}{0.308}                  \\

&  LLaMA-2 (13B)                     & \multicolumn{1}{c}{0.419}          & \multicolumn{1}{c}{0.199}           & \multicolumn{1}{c}{0.167}           & \multicolumn{1}{c}{0.089}           & 0.272          & \multicolumn{1}{c}{0.274}          & \multicolumn{1}{c}{0.271}          & \multicolumn{1}{c}{0.233}          & \multicolumn{1}{c}{0.267}                \\  

& LLaMA-3 (8B)                         & \multicolumn{1}{c}{0.573}          & \multicolumn{1}{c}{0.202}           & \multicolumn{1}{c}{0.234}             & \multicolumn{1}{c}{0.156}             & 0.336       
 & \multicolumn{1}{c}{0.356}          & \multicolumn{1}{c}{0.279}          & \multicolumn{1}{c}{0.310}          & \multicolumn{1}{c}{0.294}                 \\ 

& LLaMA-3.1 (8B)  &  0.631&  0.101&  0.307&  0.157&  0.353& 0.354&  0.375&  0.343&  0.362           \\ 

 & LLaMA-3 (70B)                        & \multicolumn{1}{c}{0.741}          & \multicolumn{1}{c}{0.192}           & \multicolumn{1}{c}{0.608}             & \multicolumn{1}{c}{0.584}             &  0.531         
&  \multicolumn{1}{c}{0.628}          & \multicolumn{1}{c}{0.295}          & \multicolumn{1}{c}{0.314}          & \multicolumn{1}{c}{{0.563}}                 \\ 

& LLaMA-3.1 (70B)     & 0.698& 0.259& 0.713& 0.665&0.627& 0.626& \underline{0.638}& \underline{0.632}& \underline{0.589}         \\ 

&  Mistral-v0.2 (7B)                      & \multicolumn{1}{c}{0.412}          & \multicolumn{1}{c}{0.152}          & \multicolumn{1}{c}{0.041}          & \multicolumn{1}{c}{0.415}           & 0.349          & \multicolumn{1}{c}{0.339}          & \multicolumn{1}{c}{0.278}          & \multicolumn{1}{c}{0.300}          & \multicolumn{1}{c}{0.271}               \\ 

& Mistral-v0.3 (7B)  & 0.571 & 0.072& 0.252& 0.134& 0.342& 0.343& 0.351& 0.336& 0.326  \\ 

& Ministral (8B)    & 0.646& 0.114& 0.405& 0.498& 0.463& 0.461& 0.475& 0.47& 0.453   \\

& Mixtral-v1.0 (8x7B)  &  0.589& 0.084& 0.356 & 0.558& 0.455& 0.456& 0.473& 0.451& 0.416\\

\textbf{Few-Shot} &   Vicuna (7B)                           & \multicolumn{1}{c}{0.578}          & \multicolumn{1}{c}{0.183}           & \multicolumn{1}{c}{0.081}           & \multicolumn{1}{c}{0.324}           & 0.325     & \multicolumn{1}{c}{0.337}          & \multicolumn{1}{c}{0.272}          & \multicolumn{1}{c}{0.354}          & \multicolumn{1}{c}{0.311}               \\

 & Vicuna (13B)                           & \multicolumn{1}{c}{0.633}          & \multicolumn{1}{c}{0.208}           & \multicolumn{1}{c}{0.383}           & \multicolumn{1}{c}{0.288}           & 0.403        & \multicolumn{1}{c}{0.427}          & \multicolumn{1}{c}{0.315}          & \multicolumn{1}{c}{0.397}          & \multicolumn{1}{c}{0.374}           \\ 

& Phi-3-small (7B)    &  0.592& 0.185& 0.301& 0.498& 0.445& 0.444& 0.455& 0.448& 0.416 \\

& Phi-3-medium (14B)   & 0.626& 0.122& 0.095& 0.447& 0.417& 0.417& 0.445& 0.415& 0.385 \\

& Gemma-2 (9B)    & 0.705&  0.390&  0.568&  0.593&  0.572&  0.571&  0.586&  0.578&  0.526  \\

& Gemma-2 (27B)    &  0.646&  0.231 & 0.67&  0.572&  0.57&  0.57&  0.585 & 0.569&  0.511 \\

& Qwen-2.5 (7B)    &  0.676&  0.198&  0.677&  0.545&  0.57&  0.572&  0.582&  0.563&  0.555 \\

& Qwen-2.5 (14B)    &  0.699&  0.257&  \textbf{0.741}&  \underline{0.676}&  0.646 & 0.680&  \textbf{0.656}&  \textbf{0.638}&  \textbf{0.608} \\

& Qwen-2.5 (72B)    & 0.721&  0.4&  \underline{0.736}&  0.503 & 0.617&  0.635&  0.608&  0.626&  0.592 \\

& {GPT-3.5-turbo}                               & \multicolumn{1}{c}{0.602}          & \multicolumn{1}{c}{0.031}           & \multicolumn{1}{c}{0.340}          & \multicolumn{1}{c}{0.604}          &       0.467    
& \multicolumn{1}{c}{0.512}          & \multicolumn{1}{c}{0.324}          & \multicolumn{1}{c}{0.384}          & \multicolumn{1}{c}{0.368}                \\ 

& GPT-4                                 & \multicolumn{1}{c}{\textbf{0.794}}          & \multicolumn{1}{c}{\textbf{0.520}}           & \multicolumn{1}{c}{{0.728}}             & \multicolumn{1}{c}{{0.653}}            &   \textbf{0.680}     &   
\multicolumn{1}{c}{\textbf{0.745}} & \multicolumn{1}{c}{{0.492}} & \multicolumn{1}{c}{{0.473}} & \multicolumn{1}{c}{{0.559}} \\ 

& GPT-4o   &   \underline{0.783}& \underline{0.507}& 0.683& 0.641& \underline{0.664}& 0.73& 0.559& 0.449& 0.529\\

& GPT-4o-mini   &  0.763& 0.435& 0.705& \textbf{0.7}& 0.657& \underline{0.741}& 0.43& 0.404& 0.588 \\

\midrule


& LLaMA-2 (7B)                          & \multicolumn{1}{c}{0.922}          & \multicolumn{1}{c}{0.897}          & \multicolumn{1}{c}{\textbf{0.944}} & \multicolumn{1}{c}{\textbf{0.933}} & 0.926       
 & \multicolumn{1}{c}{0.923}          & \multicolumn{1}{c}{0.815}          & \multicolumn{1}{c}{0.931}          & \multicolumn{1}{c}{0.921}              \\ 

& LLaMA-2 (13B)                         & \multicolumn{1}{c}{0.929}          & \multicolumn{1}{c}{0.907}          & \multicolumn{1}{c}{0.938}          & \multicolumn{1}{c}{0.923}          & 0.925         
 & \multicolumn{1}{c}{0.954}          & \multicolumn{1}{c}{0.824}          & \multicolumn{1}{c}{\textbf{0.936}} & \multicolumn{1}{c}{{0.939}}     \\ 

 & LLaMA-3 (8B)                         & \multicolumn{1}{c}{0.935}          & \multicolumn{1}{c}{0.901}          & \multicolumn{1}{c}{0.935}          & \multicolumn{1}{c}{0.928}          & 0.926         
& \multicolumn{1}{c}{0.935}          & \multicolumn{1}{c}{0.820}          & \multicolumn{1}{c}{0.930}          & \multicolumn{1}{c}{0.924}                   \\ 

\textbf{Fine-Tuing}  & LLaMA-3.1 (8B)& \textbf{0.946} &	0.919&	\underline{0.944}&	\underline{0.934}&	\underline{0.941}&	\underline{0.953}&	\underline{0.850}&	\textbf{0.939}&	\textbf{0.945} \\      

& Mistral-v0.2 (7B)                          & \multicolumn{1}{c}{0.927}          & \multicolumn{1}{c}{0.908}          & \multicolumn{1}{c}{\textbf{0.944}} & \multicolumn{1}{c}{0.849}          & 0.882        & \multicolumn{1}{c}{{0.935}} & \multicolumn{1}{c}{{0.831}} & \multicolumn{1}{c}{{0.921}} & \multicolumn{1}{c}{{0.905}} \\ 

& Mistral-v0.3 (7B) & \underline{0.944}&	\underline{0.921}&	\textbf{0.947}&	\textbf{0.935}&	\textbf{0.942}&	\textbf{0.956}&	\textbf{0.852}&	\underline{0.937}&	\underline{0.941} \\  

& Vicuna (7B)                           & \multicolumn{1}{c}{0.937}          & \multicolumn{1}{c}{0.907}          & \multicolumn{1}{c}{0.940}          & \multicolumn{1}{c}{0.906}          & 0.932       
& \multicolumn{1}{c}{\textbf{0.956}} & \multicolumn{1}{c}{{0.823}} & \multicolumn{1}{c}{\textbf{0.936}} & \multicolumn{1}{c}{{0.939}} \\  

& Vicuna (13B)                          & \multicolumn{1}{c}{{0.942}} & \multicolumn{1}{c}{\textbf{0.923}} & \multicolumn{1}{c}{0.939}          & \multicolumn{1}{c}{0.923}          & \textbf{0.933} 
& \multicolumn{1}{c}{0.950}          & \multicolumn{1}{c}{{0.847}}          & \multicolumn{1}{c}{{0.935}} & \multicolumn{1}{c}{{0.940}}           \\

\midrule
\midrule
\multicolumn{11}{c}{\textit{Existing Attribution Evaluators}} \\

& \textsc{AutoIS (11B)}                       & \multicolumn{1}{c}{0.609}          & \multicolumn{1}{c}{-}             & \multicolumn{1}{c}{-}             & \multicolumn{1}{c}{-}             & -        & \multicolumn{1}{c}{-}          & \multicolumn{1}{c}{-}          & \multicolumn{1}{c}{-}          & \multicolumn{1}{c}{-}                  \\ 

&  \textsc{AttrScore (13B)}                    & \multicolumn{1}{c}{0.687}          & \multicolumn{1}{c}{-}             & \multicolumn{1}{c}{0.523}          & \multicolumn{1}{c}{0.541}           & 0.521         
 & \multicolumn{1}{c}{0.559}          & \multicolumn{1}{c}{0.410}          & \multicolumn{1}{c}{0.432}          & \multicolumn{1}{c}{0.353}        \\ 
\bottomrule
\end{tabular}
}
\caption{Full results on CAQA. The bold and underlined results show the best and second best results respectively. Ministral behaves abnormally in the zero-shot setting and does not obey the instruction to select one of the four attribution categories.}
\label{tab:fullresults}
  \end{minipage}
\end{table*}

\section{Details of Experiments on ALCE-FineGrained}
\label{appendix:CaseOnALCE-Manual}

ALCE-FineGrained consists of 215 manually labelled samples containing 104 supportive samples, 58 partially supportive samples, 25 contradictory samples, and 28 irrelevant samples. For the few-shot setting, we select one sample for each attribution category as demonstration. For the fine-tuning setting, we employ GPT-4 to annotate 800 samples from the ALCE benchmark as the training set. Since there are fewer contradictory and irrelevant attribution categories in the ALCE benchmark, we use GPT-4 to edit the evidence to construct contradictory and irrelevant samples, thus ensuring a balanced number of the four categories.

Table \ref{tab:ACLE-mannual} presents two ALCE-FineGrained examples, illustrating the attribution categories \textit{partially supportive} and \textit{irrelevant}, respectively. It shows that these two categories, which are not included in the previous attribution categories, are common and different in practical situations. In example 1, where the attribution category is \textit{partially supportive}, most of the answer statement (highlighted in green) is mentioned in the citation, but the key information ``\textit{The Maryland Transportation Authority}'' (highlighted in yellow) is not mentioned in the citation. This demonstrates the subtleties that can render an attribution insufficient. In example 2, which is categorised as \textit{irrelevant}, the entirety of the answer statement is irrelevant to the citation. This exemplifies a clear case of irrelevant attribution.
Notably, previous evaluators, AutoIS and AttrScore, are unable to accurately classify these cases. In contrast, Vicuna, an evaluator trained with our CAQA benchmark, successfully identifies the correct attribution categories. This underscores the effectiveness and practicality of employing the CAQA benchmark for developing attribution evaluators.

\begin{table*}[!t]
\centering
\small
\caption{Examples of the four complexity types. Reasoning Graphs show the reasoning relationships between citations-answers. \textcolor{deepgreen}{Green} represents content associated with the answer, \textcolor{gray}{gray} indicates excluded content, and \textcolor{orange}{orange} indicates the common term connecting the citations.
}
\label{tab:examples_of_complexity}
\scalebox{0.85}{
\begin{tabularx}{1.1\textwidth}{lXl}
\toprule
\textbf{Complexity} & \textbf{Examples} & \textbf{Reasoning Graphs}  \\
\midrule
\raisebox{-2.9\height}{\textbf{Single}} & \textbf{Question:} Which radio program episode appears in All Things Considered?  \newline \textbf{Answer:} The radio program episode in which All Things Considered appears is Remorse: The 14 Stories of Eric Morse [1].
\newline \textbf{Citations:} [1] \textcolor{deepgreen}{\textbf{\textit{Remorse: The 14 Stories of Eric Morse is an episode of the radio program All Things Considered}}}....
& \raisebox{-1.1\height}{ \includegraphics[scale=0.15]{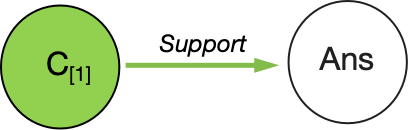}}
\\
\midrule
\raisebox{-3.2\height}{\textbf{Union}} & \textbf{Question:} Which university did Rick Scott attend?
\newline \textbf{Answer:} Rick Scott attended the University of Missouri–Kansas City and Southern Methodist University [1][2].
\newline \textbf{Citations:} [1] \textcolor{deepgreen}{\textbf{\textit{Rick Scott graduated from the University of Missouri–Kansas City }}} ... [2] \textcolor{deepgreen}{\textbf{\textit{Rick Scott earned a juris doctor degree by working his way through Southern Methodist University}}}, ...   
& \raisebox{-1.2\height}{ \includegraphics[scale=0.15]{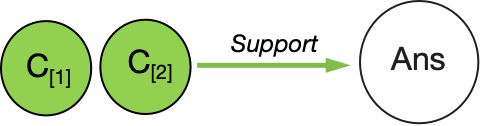}}
\\
\midrule
\raisebox{-4.8\height}{\textbf{Intersection}}  & \textbf{Question:} The computer designer for Macintosh 128k and NeXT computer was whom?
\newline \textbf{Answer:} {{{The computer designer for Macintosh 128k and NeXT computer was Steve Jobs}}} [1][2].
\newline \textbf{Citations:} [1] \textcolor{deepgreen}{\textbf{\textit{The computer designer for Macintosh 128k was}}} \textcolor{gray}{\textbf{\textit{Jerry Manock}}},  \textcolor{deepgreen}{\textbf{\textit{ who worked with Steve Jobs to develop the vertical body}}} ... [2] ...\textcolor{deepgreen}{\textbf{\textit{Several former Apple employees followed Jobs to NeXT}}}, \textcolor{gray}{\textbf{\textit{including Joanna Hoffman, Bud Tribble, George Crow, Rich Page}}}...
& \raisebox{-2.2\height}{\includegraphics[scale=0.15]{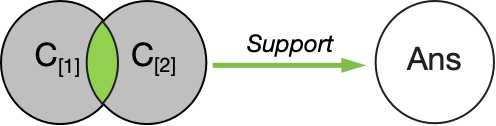}}
\\
\midrule
\raisebox{-4.9\height}{\textbf{Concatenation}} & \textbf{Question:} What are the official languages in the politician Mohammad Najibullah's country?
\newline \textbf{Answer:} {{{Pashto and Dari are the official languages in the politician Mohammad Najibullah's country.}}} [1][2].
\newline \textbf{Citations:} [1] \textcolor{deepgreen}{\textbf{\textit{Mohammad Najibullah was the president of}}} \textcolor{orange}{\textbf{\textit{Afghanistan}}} \textcolor{deepgreen}{\textbf{\textit{from 1986 to 1992}}} ... [2] \textcolor{orange}{\textbf{\textit{Afghanistan}}} \textcolor{deepgreen}{\textbf{\textit{s a multilingual country, where Pashto and Dari (a dialect of Persian) are the official languages with}}} ...
& \raisebox{-1.9\height}{ \includegraphics[scale=0.15]{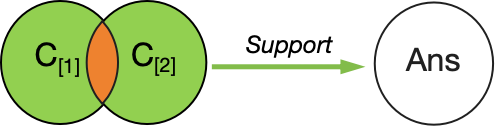}}
\\

\bottomrule
\end{tabularx}
}

\vspace{-0.1cm}
\end{table*}

\begin{table*}[h]
\small
\begin{tcolorbox}

\textbf{Example 1}

\textbf{Question:} {The Maryland Transportation Authority is in charge of what landmarks?} 

\textbf{Answer Statement:} \textcolor{deeperyellow}{The Maryland Transportation Authority} is \textcolor{deepgreen}{in charge of providing funding for the construction of revenue-producing transportation facilities for the Maryland Department of Transportation, which have included improvements at the Port of Baltimore and the Baltimore-Washington International Airport}

\textbf{Citation:} of a commercial enterprise. Its capital projects and operations are funded by tolls, concessions, investment income, and revenue bonds. In addition to its own toll facilities, the Authority \textcolor{deepgreen}{finances construction of other revenue-producing transportation facilities for the Maryland Department of Transportation (MDOT). These projects have included improvements at the Port of Baltimore and the Baltimore-Washington International Airport.} To provide construction funding, the Authority issues revenue bonds, which will be paid off over a period of years by tolls and other user fees generated by the facilities. The MDTA can issue either taxable or exempt bonds to finance large scale projects.

\textbf{AutoIS:} \textcolor{blue}{Supportive} \textcolor{red}{\ding{55}}

\textbf{AttrScore:} \textcolor{blue}{Irrelevant} \textcolor{red}{\ding{55}}

\textbf{Vicuna$^\dagger$:} \textcolor{blue}{Partially Supportive} \textcolor{green}{\checkmark}

\rule{\linewidth}{0.2mm}

\textbf{Example 2}

\textbf{Question:} {When did the last season of jersey shore air?} 

\textbf{Answer Statement:} {\textcolor{deeperyellow}{The TV show Jersey Shore aired its final episode on December 20, 2012.}}

\textbf{Citation:} 8.56 million viewers, only to set another record with the airing of the fourth episode, which garnered 8.87 million viewers. On January 25, 2011, it was confirmed that the show had been renewed for a fourth season, to be filmed in Italy during the first half of 2011. The fourth season premiered August 4, 2011. MTV confirmed in June 2011 that the fifth season would return to Seaside Heights. Believed complications caused by Nicole Polizzi's pregnancy, and several cast members (including Polizzi, DelVecchio, and Farley) receiving spin-offs sparked talk about the future of the series past the fifth season, however

\textbf{AutoIS:} \textcolor{blue}{Supportive} \textcolor{red}{\ding{55}}

\textbf{AttrScore:} \textcolor{blue}{Contradictory} \textcolor{red}{\ding{55}}

\textbf{Vicuna$^\dagger$:} \textcolor{blue}{Irrelevant} \textcolor{green}{\checkmark}

\end{tcolorbox}
\vspace{-2mm}
\caption{Two examples of the results of the three attribution evaluators on ALCE-FineGrained. Content in yellow highlights portions of the answer statement not found in the citation, whereas green indicates content present in the citation.}
\label{tab:ACLE-mannual}
\end{table*}

\section{Analysis of Existing Attributed QA Systems}
\label{appendix:AnalysiOfAQASytems}
Following the work of Gao et al. \citep{gao2023enabling} we reproduce the attributed question answering system based on Vicuna-13B model, noted for its effectiveness in smaller language model configurations. Specifically, we provide the model with the top-3 retrieved passages and instruct the model to cite them accordingly. The retrieved passages and the instruction are consistent with the original implementation. Upon reviewing 234 instances of the system, our analysis revealed that: 44.4\% of the instances accurately cited evidence supporting their answers, while 24.8\% cited evidence that only partially supported the answers. Contradictory evidence was cited in 10.7\% of cases, and 12.0\% of the responses involved citations of irrelevant evidence. Additionally, 8.1\% of the cases were categorized under other issues, including incomplete or unclear answers.
The predominant challenges in incorrect attributions are identified as \textit{partially supportive}, \textit{contradictory}, and \textit{irrelevant} citations, with \textit{partially supportive} citations being the most common problem.

\section{Human Annotation}
\label{appendix:HumanAnnotation}

\begin{figure*}[h]
  \centering
  \includegraphics[width=0.95\textwidth]{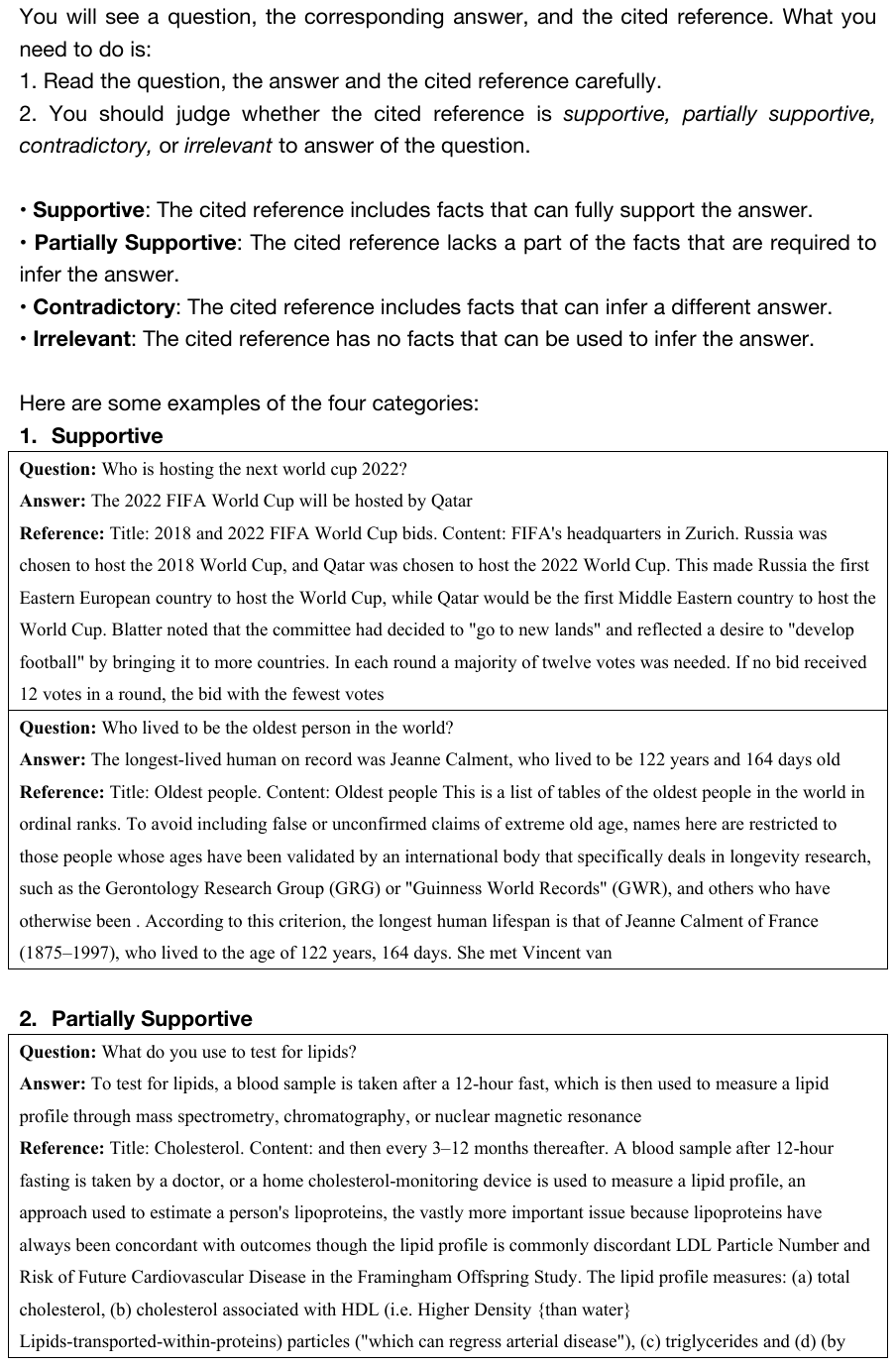}
  \caption{First page of the annotation guidelines.}
  \label{fig:annotationguidelines1}
\end{figure*}

\begin{figure*}[h]
  \centering
  \includegraphics[width=0.95\textwidth]{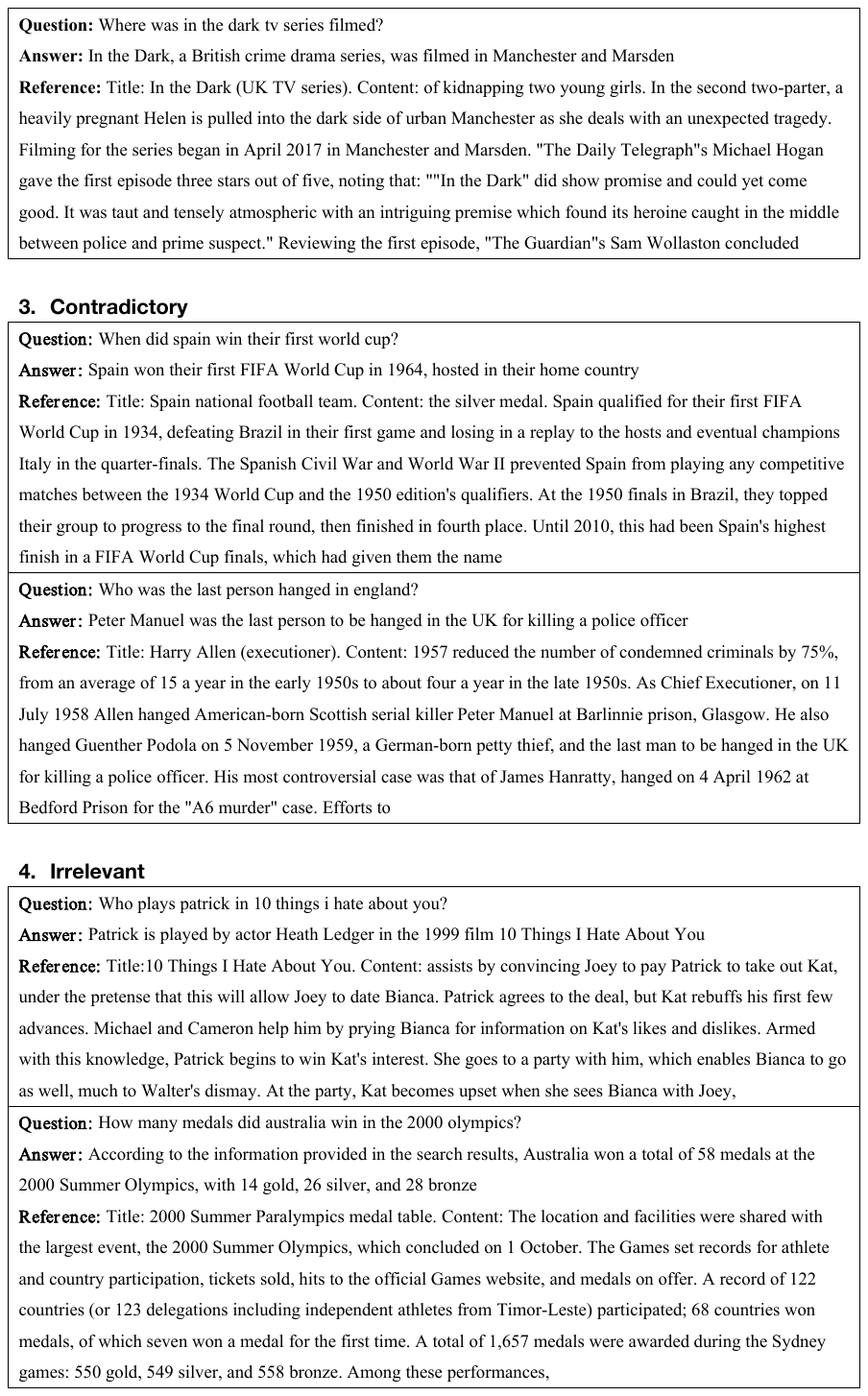}
  \caption{Second page of the annotation guidelines.}
  \label{fig:annotationguidelines2}
\end{figure*}

The human annotation process for our study was conducted by the authors themselves, eliminating the need for external paid services. Three of our annotators were asked to read these guidelines carefully. Only annotators with a thorough understanding of the guidelines and the task were allowed to participate in the manual evaluation. We ensured the reliability of the results by retaining only those annotations that were aligned across all three annotators.  Annotation guidelines are shown in Fig. \ref{fig:annotationguidelines1} and \ref{fig:annotationguidelines2}.

For the CAQA benchmark, we sampled 400 cases and provided them to three annotators. Among these, 2 examples were deemed unclear by more than one annotator and subsequently discarded. This resulted in 398 valid annotated examples. The Fleiss' Kappa score for these annotations was 0.780, reflecting a substantial level of agreement. For the ALCE-FineGrained benchmark, we sampled 300 cases and provided them to three annotators. Of these, 24 examples were deemed unclear and were excluded, leaving 276 valid annotated examples. The Fleiss' Kappa score for these annotations was 0.748, also indicating a substantial level of agreement.

Additionally, we analyzed the labeling process and identified patterns of agreement and disagreement among annotators. Overall, the category with the highest agreement was ‘supportive’, followed by ‘irrelevant’, while the categories ‘partially supportive’ and ‘contradictory’ exhibited lower agreement. Specifically, annotators often confused ‘partially supportive’ with ‘irrelevant’, frequently misclassifying examples between these two categories. This confusion arose because distinguishing between them required identifying sub-facts (i.e., the smallest semantic units containing a subject, verb, and object) from answer statements and evidence. The process often required domain-specific knowledge, which annotators lacked, leading them to rely on co-occurring keywords rather than deeper semantic understanding. Similarly, annotators had difficulty identifying ‘contradictory’ due to a lack of domain knowledge. This led to misclassifications of ‘contradictory’ as either ‘partially supportive’ or ‘irrelevant’. Determining contradictions requires nuanced reasoning beyond surface-level overlaps, which posed challenges in the absence of sufficient domain context.

\section{License}

Our dataset will be distributed under the CC BY-SA 4.0 license.

\end{document}